\documentclass{article}

\usepackage[final,nonatbib]{neurips_2021}

\usepackage[utf8]{inputenc} 
\usepackage[T1]{fontenc}    
\usepackage{hyperref}       
\usepackage{url}            
\usepackage{booktabs}       
\usepackage{amsfonts}       
\usepackage{nicefrac}       
\usepackage{microtype}      
\usepackage{xcolor}         
\usepackage{enumitem}
\usepackage{graphicx}
\usepackage{amsmath,amsthm}
\newtheorem{definition}{Definition}[section]
\newtheorem{theorem}{Theorem}
\newtheorem{proposition}{Proposition}
\newtheorem{corollary}{Corollary}
\newtheorem{lemma}{Lemma}

\newenvironment{customproposition}[1]
  {\renewcommand\theproposition{#1}\proposition}
  {\endproposition}
  
\newenvironment{customlemma}[1]
  {\renewcommand\thelemma{#1}\lemma}
  {\endlemma}

\usepackage{bm}
\usepackage{xspace}
\usepackage{tikz}
\usetikzlibrary{bayesnet}
\usetikzlibrary{arrows,shadows,shapes,backgrounds,decorations,snakes,fit}
\usetikzlibrary{fit,positioning}
\tikzset{fit margins/.style={/tikz/afit/.cd,#1,
    /tikz/.cd,
    inner xsep=\pgfkeysvalueof{/tikz/afit/left}+\pgfkeysvalueof{/tikz/afit/right},
    inner ysep=\pgfkeysvalueof{/tikz/afit/top}+\pgfkeysvalueof{/tikz/afit/bottom},
    xshift=-\pgfkeysvalueof{/tikz/afit/left}+\pgfkeysvalueof{/tikz/afit/right},
    yshift=-\pgfkeysvalueof{/tikz/afit/bottom}+\pgfkeysvalueof{/tikz/afit/top}},
    afit/.cd,left/.initial=2pt,right/.initial=2pt,bottom/.initial=2pt,top/.initial=2pt}

\usepackage{bm}
\usepackage{color}
\usepackage{amssymb}

\usepackage{algorithmic}
\usepackage{algorithm}

\usepackage{multirow}

\DeclareMathOperator*{\argmax}{arg\,max}

\newcommand{\cz}[1]{{\color{black} #1}} 

\newcommand{\name}[0]{GINA\xspace}

\usepackage{wrapfig}
\usepackage{arydshln}

\title{Identifiable Generative Models for \\ Missing Not at Random Data Imputation}

\author{
Chao Ma\textsuperscript{1,2} \thanks{This work was performed when the authors were (part-time) associated with Microsoft Research, Cambridge}
\quad 
Cheng Zhang\textsuperscript{2}
\\\
$^1$University of Cambridge \quad $^2$Microsoft Research Cambridge
\\\
\texttt{cm905@cam.ac.uk}
\\
\texttt{cheng.zhang@microsoft.com}
}

\begin{document}

\maketitle

\begin{abstract}

Real-world datasets often have missing values associated with complex generative processes, where the cause of the missingness may not be fully observed. This is known as missing not at random (MNAR) data. However, many imputation methods do not take into account the missingness mechanism, resulting in biased imputation values when MNAR data is present. Although there are a few methods that have considered the MNAR scenario, their model's identifiability under MNAR is generally not guaranteed. That is, model parameters can not be uniquely determined even with infinite data samples, hence the imputation results given by such models can still be biased. This issue is especially overlooked by many modern deep generative models. In this work, we fill in this gap by systematically analyzing the identifiability of generative models under MNAR. Furthermore, we propose a practical deep generative model which can provide identifiability guarantees under mild assumptions, for a wide range of MNAR mechanisms. Our method demonstrates a clear advantage for tasks on both synthetic data and multiple real-world scenarios with MNAR data.

\end{abstract}

\section{Introduction} \label{sec:intro}

Missing data is an obstacle in many data analysis problems, which may seriously compromise the performance of machine learning models, as well as downstream tasks based on these models. Being able to successfully recover/impute missing data in an unbiased way is the key to understanding the structure of real-world data. This requires us to identify the underlying data-generating process, as well as the probabilistic mechanism that decides which data is missing.

\begin{figure}
    \centering
    \includegraphics[width=0.75\textwidth]{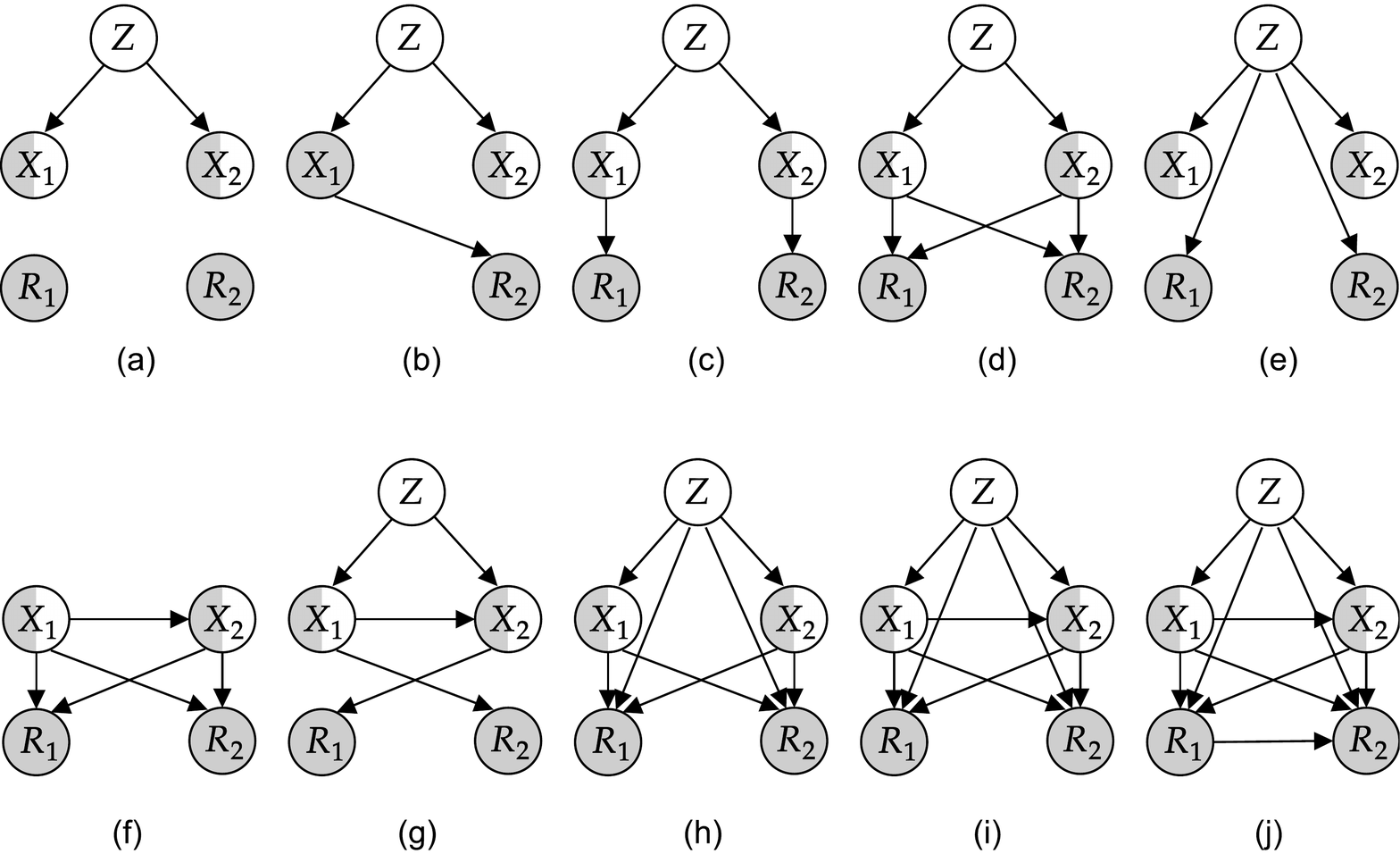}
    \caption{Exemplar missing data situations. 
    \textbf{(a)}: MCAR; \textbf{(b)}: MAR; \textbf{(c)-(i)}: MNAR.  
    }
    \label{fig:MNAR2}
\end{figure}

In general, there are three types of missing mechanisms \cite{rubin1976inference}. The first type 
is missing completely at random (MCAR), where the probability of a data entry being missing is independent of both the observed and unobserved data (Figure \ref{fig:MNAR2} (a)). In this case, no statistical bias is introduced by MCAR.
The second type is missing at random (MAR), which assumes that the missing data mechanism is independent of the value of unobserved data (Figure \ref{fig:MNAR2} (b)). Under this assumption,  maximum likelihood learning methods without explicit modeling of the missingness mechanism can be applied by marginalizing out the missing variables \cite{em_dempster1977maximum,ma2018partial,ma2018eddi}.  
However, both MCAR and MAR do not hold in many real-world applications, 
such as recommender systems \cite{hernandez2014probabilistic,jannach2010recommender}, 
healthcare \cite{jakobsen2017and}, and surveys \cite{shrive2006dealing}. For example, in a survey, participants with financial difficulties are more likely to refuse to complete the survey about financial incomes. This is an example of missing not at random (MNAR),
where the cause of the missingness (financial income) can be unobserved. In this case, ignoring the missingness mechanism will result in biased imputation, which will jeopardize down-stream tasks.  

There are few works considering the MNAR setting in scalable missing value imputation. On the one hand, many practical methods for MNAR does not have identifiability guarantees \cite{notmiwae, hernandez2014probabilistic, little2019statistical}. That is, the parameters can not be uniquely determined, even with access to infinite samples \cite{miao2016identifiability, rothenberg1971identification}. As a result, missing value imputation based on such parameter estimation could be biased.   On the other hand, there are theoretical analyses on the identifiability in certain scenarios \cite{miao2016identifiability, miao2015identification, miao2018identification, mohan2013graphical, sportisse2020estimation,  tang2003analysis, wang2014instrumental}, but without associated practical algorithms for flexible and scalable settings (such as deep generative models). Moreover, MNAR data have many possible cases (Figure \ref{fig:MNAR2}) based on different independence assumptions \cite{mohan2013graphical}, making the discussion of identifiability difficult. 
This motivates us to fill this gap by extending identifiability results of deep generative models to different missing mechanisms, and provide a scalable practical solution.
 Our contribution are threefold: 
\begin{itemize} [leftmargin=1em,labelwidth=*,align=left]
    \item We provide a theoretical analysis of identifiability for generative models under different MNAR scenarios (Section \ref{sec:theoretical}).
    More specifically, we provide sufficient conditions, under which the ground truth parameters can be uniquely identified via maximum likelihood (ML) learning using observed information \cite{little2019statistical}. We also demonstrate how the assumptions can be relaxed in the face of real-world datasets. This provides foundation for practical solutions using deep generative models.
    \item Based on our analysis, we propose a practical algorithm model based on VAEs (Section \ref{sec:practical_model}), named GINA (deep \underline{g}enerative \underline{i}mputation model for missing \underline{n}ot \underline{a}t random). This enables us to apply flexible deep generative models in a principled way, even in the presence of MNAR data.  
    \item  We demonstrate the effectiveness and validity of our approach by experimental evaluations (Section 
    \ref{sec:exp}) on both synthetic data modeling, missing data imputation in real-world datasets, as well as downstream tasks such as active feature selection under missing data. 
    \end{itemize}

\section{Backgrounds} \label{sec:background}
\subsection{Problem \cz{Setting}} \label{sec:backgrounds_1}

\cz{A critical component to develop model to impute MNAR data is the model identifiablity \cite{koller2009probabilistic, rothenberg1971identification}.  We give the definition below:}
\begin{definition}[Model identifiability]\label{def:idf} Assume $p_\theta(X)$ is a distribution of some random variable $X$, $\theta$ is its parameter that takes values in some parameter space $\Omega_\theta$. Then, if $p_\theta(X)$ satisfies $
p_{\theta_1}(X)\neq p_{\theta_2}(X) \Longleftrightarrow \theta_1 \neq \theta_2, \forall \theta_1,\theta_2 \in \Omega_\theta$, 
we say that $p_\theta$ is identifiable w.r.t. $\theta$ on $\Omega_\theta$.
\end{definition}
In other words, a model $p_\theta(X)$ is identifiable, if different parameter configurations implies a different probabilistic
distributions over the observed variables. With identifiability guarantee, if the model assumption is correct, the true generation process can be recovered.
Next, we introduce necessary notations and of missing data, and set up a concrete problem setting.

\noindent\textbf{Basic Notation.~}  Similar to the notations introduced by \cite{notmiwae, rubin1976inference}, let $X$ be the complete set of variables in the system of interest. We call it \emph{observable variables}. Let $\mathcal{I} = \{1,...,D\}$ be the \emph{index set} of all observable variables, i.e., $X = \{X_i|i\in \mathcal{I}\}$. Let $X_o$ denote the set of actually \emph{observed variables}, here $O \in \mathcal{I}$ is a index set such that $X_o \subset X$. We call $O$ the \emph{observable pattern}. Similarly, $X_u$ denotes the set of \emph{missing/unobserved variables}, and $X = X_o \bigcup X_u$. Additionally, we use $R$ to denote the missing mask indicator variable, such that $R_i = 1$ indicates $X_i$ is observed, and $R_i = 0$ indicates otherwise.   We call a probabilistic distribution $p(X)$ on $X$ the \emph{reference distribution}, that is, the distribution that we would have observed if no missing mechanism is present; and we call the conditional distribution $p(R|X)$ the \emph{missing mechanism}, which decides the probability of each $X_i$ being missing.
%
%
Then, we can define the marginal distribution of \emph{partially observed variables}, which is given by $\log p(X_o, R) = \log \int_{X_u} p(X_o, X_u, R) d X_u$ .  Finally, we will use lowercase letters to denote the \emph{realized values} of the corresponding random variable. For example, $(x_o,r) \sim p(X_o, R)$ is the realization/samples of $X_o$ and $R$, and the dimensionality of $x_o$ may vary for each realizations.

\noindent\textbf{Problem setting.~} Suppose that we have a ground truth data generating process, denoted by $ p_{\mathcal{D}}(X_o,R)$, from which we can obtain (partially observed) samples $(x_o, r) \sim p_{\mathcal{D}}(X_o, R)$. We also have a model to be optimized, denoted by $p_{(\theta,\lambda)}(X_o, X_u, R)$,where $\theta$ is the parameter of reference distribution $p_\theta(X)$, and $\lambda$ the parameter of missing mechanism $p_\lambda(R|X)$. Our goal can then be described as follows:
\begin{itemize} [nosep,leftmargin=1em,labelwidth=*,align=left]
\item To establish the identifiability of the model $p_{(\theta,\lambda)}(X_o,R)$. That is, we wish to uniquely and correctly identify $\hat{\theta}$, such that $p_{\hat{\theta}}(X) = p_{\mathcal{D}}(X)$, given infinite amount of partially observed data samples from ground truth, $(x_o, r) \sim p_{\mathcal{D}}(X_o, R)$.
\item Then, given the identified parameter, we will be able to perform missing data imputation, using $p_{\hat{\theta}}(X_u|X_o)$. If our parameter estimate is unbiased, then our imputation is also unbiased, that is, $p_{\hat{\theta}}(X_u|X_o) = p_{\mathcal{D}}(X_u|X_o)$ for all possible configurations of $X_o$.
\end{itemize}

\subsection{Challenges in MNAR imputation}  \label{sec:backgrounds_2}
\label{sec:rubin} 
Recall the three types of missing mechanisms: if data is MCAR, $p(R|X) = p(R)$; if if data is MAR, $p(R|X) = p(R|X_o)$; otherwise, we call it MNAR. When missing data is MCAR or MAR, missing mechanism can be ignored when performing maximum likelihood (ML) inference based only on the observed data \cite{rubin1976inference}, as:
\begin{equation*}
    \arg \max_\theta \mathbb{E}_{(x_o,r) \sim p_{\mathcal{D}}(X_o,R)} \log p_\theta(X_o = x_o) = \arg \max_\theta \mathbb{E}_{(x_o,r)\sim p_{\mathcal{D}} (X_o,R)} \log p_\theta (X_o = x_o,R = r) 
\end{equation*}
where $\log p(X_o) = \log \int_{X_u} p(X_o, X_u) d X_u$.
In practice, ML learning on $X_o$ can done by EM algorithm \cite{em_dempster1977maximum,little2019statistical}. 
However, when missing
data is MNAR, the above argument does not hold, and the missing data mechanism cannot be ignored during learning. Consider the representative graphical model example in Figure \ref{fig:MNAR2} (d), which has appeared in many context of machine learning. In this graphical model, $X$ is the cause of $R$, and the connections between $X$ and $R$ are fully connected, i.e., each single node in $R$ are caused by the entire set $X$. All nodes in $R$ are conditionally independent from each other given $X$. 

Clearly, this is an example of a data generating process with MNAR mechanism.  In this case, Rubin proposed to jointly optimize both the reference distribution $p_\theta(X)$ and the missing data mechanism $p_\lambda(R|X)$, by maximizing:
\begin{equation}\arg \max_{\theta, \lambda}\mathbb{E}_{(x_o,r)\sim p_{\mathcal{D}} (X_o,R)} \left[ \log p_{\theta}(X_o = x_o) + \log p_\lambda(R = r|X_o = x_o) \right]
    \label{eq:rubin_mnar}
\end{equation}
 This factorization is referred as \emph{selection modeling} \cite{notmiwae,little2019statistical}. 
There are multiple challenges if we want to Eq. \ref{eq:rubin_mnar} to obtain a practical model that provide unbiased imputation. First, we need model assumption to be consistent with the real-world data generation process,  $p_{\mathcal{D}}(X_o,R)$. Given a wide range of possible MNAR scenarios, it is a challenge to design a general model. Secondly, the model need to be identifiable to enable the possibility to learn the underlying process which leads to unbiased imputation.

\subsection{Variational Autoencoders and its identifiability} \label{sec:idf_vae} Variational auto-encdoers \cite{kingma2013auto,rezende2014stochastic,zhang2018advances} is a flexible deep generative model that is commonly used for estimating densities of $p_\mathcal{D}(X)$. It takes the following form:
\begin{equation}
    \log p_{\theta}(X) =  \log \int_{Z} dZ p_{\theta}(X | Z) p(Z), 
\end{equation}
where $Z$ is some latent variable model with prior $p(Z)$, and $p_{\theta}(X | Z)$ is given by $p_{\theta}(X | Z) = \mathcal{N}(f_\theta(Z), \sigma)$, with $f_\theta(\cdot)$ being a neural network parameterized by $\theta$. Generally, VAEs do not have identifiability guarantees w.r.t. $\theta$ \cite{khemakhem2020variational}.  Nevertheless, \cz{inspired by the identifiablity of nonlinear ICA,} \cite{khemakhem2020variational} shows that the identifiability of VAE can be established \cz{up to equivalence permutation under mild assumptions}, if the unconditional prior $p(Z)$ of VAE is replaced by the following the conditionally factorial exponentially family prior, 
\begin{equation}
    p_{T,\zeta}(Z|U) \propto \prod_{i=1} Q(Z_i)\exp[ \sum_{j=1}^KT_{i,j}(Z_i)\zeta_{i,j}(U) ],
\end{equation}
 where $U$ is some additional observations (called auxiliary variables), $Q(Z_i)$ is some base measure,  $\mathbf{T}_i(U) = ( T_{i,1}, ..., T_{i,K} )$ the sufficient statistics, and $\bm{\zeta}_i(U) = ( \zeta_{i,1}, ..., \zeta_{i,K} )$ the corresponding natural parameters. Then, the new VAE model  given by
 \begin{equation}
    \log p_{\theta}(X|U) =  \log \int_{Z} dZ p_{\theta}(X | Z) p_{T,\zeta}(Z|U) \label{eq:identifi_VAE}
\end{equation}
is identifiable (Theorem 1 and 2 of \cite{khemakhem2020variational}, see Appendix \ref{app:ivae}).
We call the model (\ref{eq:identifi_VAE}) the \emph{identifiable VAE}. Unfortunately, this identifiability results for VAE only hold when all variables are fully observed; thus, it cannot be immediately applied to address the challenges of dealing with MNAR data stated in Section \ref{sec:backgrounds_2}. Next, we will analyze the identifiablity of generative models under general MNAR settings (Section \ref{sec:theoretical}), and propose a practical method that can be used in MNAR  (Section \ref{sec:practical_model}).

 \section{Establishing model identifiability under MNAR} \label{sec:theoretical}
One key issue of training probabilistic models under MNAR missing data is its identifiability. 
%
%
Recall that (Definition \ref{def:idf}) model identifiability characterize the property that the mapping from parameter $\theta$ to the distribution $p_\theta(X)$ is one-to-one. This is often closely related to maximum likelihood learning. In fact, it is not hard to show that Definition \ref{def:idf} is equivalent to the following Definition \ref{def:idf2}:
\begin{definition}[Equivalent definition of identifiability] \label{def:idf2}
We say a model $p_\theta(X)$ is identifiable, if:
\begin{equation} 
    \arg \max_{\theta \in \Omega_\theta} \mathbb{E}_{x\sim p_{\theta^*}(X)} \log p_\theta(X = x) = \theta^*,\ \  \forall \theta^* \in \Omega_\theta \label{eq:complete_ML}
\end{equation}
\end{definition}

 In other words, the ``correct'' model parameter $\theta^*$ can be identified via maximum likelihood learning (under complete data), and the ML solution is unbiased. Similarly, when MNAR missing mechanism is present, we perform maximum likelihood learning on both $X_o$ and $R$ using Eq. \ref{eq:rubin_mnar}.
 Thus, we need $\log p_{\theta,\lambda}(X_o,R)$ to be identifiable under MNAR, so that we can correctly identify the ground truth data generating process, and achieve unbiased imputation. 
 The identifiability of  $\log p_{\theta,\lambda}(X_o,R)$ under MNAR is usually not guaranteed, even in some simplistic settings \cite{ miao2016identifiability}. In this section, we will give the sufficient conditions for model identifiability under MNAR, and study how these can be relaxed for real-world applications

\subsection{Sufficient conditions for  identifiability under MNAR}\label{sec:theory_for_h}

In this section, we give sufficient conditions where the model parameters $\theta$ can be uniquely identified by Rubin's objective, Eq. \ref{eq:rubin_mnar}. Our aim is to i), find a set of model assumptions, so that it can cover many common scenarios and be flexible for practical interests; and ii), under those conditions, we want to show that its parameters can be uniquely determined by the partial ML solution Eq. \ref{eq:rubin_mnar}. As shown in Figure \ref{fig:MNAR2}, MNAR have many possible difference cases depending on its graphical structures. We want our results to cover every situation. 

Instead of doing case by case analysis, we will start our identifiability anaylsis with one fairly general case as the example shown in Figure \ref{fig:MNAR2} (h) where the missingness can be caused by other partially observed variable, by itself (self-masking) or by latent variables. Then, we will discuss how these analysis can be applied to other MNAR scenarios in Section \ref{sec:relax}.

\noindent\textbf{Data setting D1.} Suppose the ground truth data generation process 
 satisfies the following conditions: all variables $X$ are generated from a shared latent confounder $Z$, and there are no connections among $X$; and 
 the missingness indicator $R$ variable can not be the parent of other variables. A typical example of such distribution is depicted in Figure \ref{fig:MNAR2} (h).
  We further assume that $p_{\mathcal{D}}(X_o, X_u, R)$ has the following parametric form: $p_{\mathcal{D}}(X_o, X_u, R) = \int_Z \prod_d p_{\theta^*_d}(X_d|Z) p(Z) p_{\lambda^*}(R|X,Z)dZ$, where 
 $p_{\lambda^*}(R|X,Z) = \prod_d p_{\lambda^*_d}(R_d|X,Z)$, for some $\theta^*, \lambda^*$.

Then, consider the following model:

\noindent\textbf{Model assumption A1.} We assume that our model has the same graphical representation,  as well as parametric form as \textbf{data setting D1}, that is, our model can be written as:
\begin{equation}
    p_{\theta,\lambda}(X_o,R) = \int_{X_u,Z} dX_u dZ \prod_d p_{\theta_d}(X_d|Z) \prod_d p_{\lambda_d}(R_d|X,Z)  p(Z)
    \label{eq:theoretical_model}
\end{equation} 
Here, $(\theta,\lambda) \in \Omega$ are learnable parameters that belong to some parameter space $\Omega = \Omega_\theta \times \Omega_\lambda$. Each $\theta$ is the parameter that parameterizes the conditional distribution that connects $X_d$ and $Z$, $p_{\theta_d}(X_d|Z)$. Assume that the ground truth parameter of $p_{\mathcal{D}}$ belongs to the model parameter space, $(\theta^*, \lambda^*) \in \Omega$.

Given such a model, our goal is to correctly identify the ground truth parameter settings given partially observed samples from $p_{\mathcal{D}}(X_o,X_u,R)$.
That is, let $(\hat{\theta},\hat{\lambda}) = \arg \max_{(\theta, \lambda)\in \Omega} \mathrm{E}_{(x_o,r) \sim p_{\mathcal{D}}(X_o,R)} \log p_{(\theta,\lambda)}(X_o = x_o,R = r)$, we would like to achieve $\hat{\theta} = \theta^*$. In order to achieve this, we must make additional assumptions. 

\noindent\textbf{Assumption A2.} Subset identifiability: There exist a partition\footnote{It can be arbitrary partition in the set theory sense. \label{note1} } of $\mathcal{I}$, denoted by $\mathcal{A}_\mathcal{I} = \{O_s\}_{1\leq s \leq S}$, such that: for all $O_s \in \mathcal{A}_\mathcal{I}$, $p_\theta(X_{o_s})$ is identifiable on a subset of parameters $\{\theta_d|d \in O_s \}$.

This assumption basically formalizes the idea of divide and conquer: we partition the whole index set into several smaller subsets $\{O_s\}_{1\leq s \leq S}$, on which each reference distribution $p_\theta(X_{O_s})$ is only responsible for the identifiability on a subset of parameters. 


\noindent\textbf{Assumption A3.} There exists a collection of observable patterns, denote by $\Bar{\mathcal{A}_\mathcal{I}} := \{O'_l\}_{1\leq l \leq L}$, such that: 1), $\Bar{\mathcal{A}_\mathcal{I}}$ is a cover \textsuperscript{\ref{note1}} of $\mathcal{I}$; 2), $p_{\mathcal{D}}(X, R_{O'_l} = 1, R_{\mathcal{I}\setminus {O'_l} })>0$ for all $1\leq l \leq L$; and 3), for all index $c\in O'_l$, there exists $O_s \in \mathcal{\mathcal{A}_\mathcal{I}}$ defined in \textbf{A2}, such that $c \in O_s \subset O'_l$.

This assumption is about the strict positivity of the ground truth data generating process, $p_{\mathcal{D}}(X_o,X_u,R)$. Instead of assuming that complete case data are available as in \cite{mohan2013graphical}, here we assumes we should at least have some observations, $p_{\mathcal{D}}(X,R_o =1, R_u = 0)>0$ for $O \in \hat{\mathcal{A}}_\mathcal{I}$, on which $p_\theta(X_o)$ is identifiable.

To summarize,  \textbf{A1} ensures that our model has the same graphical representation/parametric forms as the ground truth; \textbf{A2} $p_\theta(X_o) = \int_{X_u} p_\theta(X_o,X_u) dX_u$ should be at least identifiable for a collection of observable patterns that forms a partition of $\mathcal{I}$; and \textbf{Assumption A3} ensures that $p_{\mathcal{D}}(X_o,X_u, R)$ should be positive for certain \emph{important} patterns (i.e., those on which $p_\theta(X_o)$ is identifiable). In Appendix \ref{app:proof1}, we will provide a practical example that satisfies those assumptions. Given these assumptions, we have the following proposition (See Appendix \ref{app:proof1} for proof.):

\vspace{5pt}
\begin{proposition}
 [Sufficient conditions for identifiability under MNAR] \label{prop:1} Let $p_{\theta,\lambda}(X_o,X_u,R)$ be a model on the observable variables $X$, and missing pattern $R$, and $p_{\mathcal{D}}(X_o,X_u,R)$ be the ground truth distribution. Assume that they satisfies \textbf{Data setting D1}, \textbf{Assumptions A1, A2} and \textbf{A3}. 
 
Let $\Theta = \arg \max_{(\theta, \lambda)\in \Omega} \mathrm{E}_{(x_o,r) \sim p_{\mathcal{D}}(X_o,R)} \log p_{(\theta,\lambda)}(X_o = x_o,R = r)$ be the set of ML solutions of Equation \ref{eq:rubin_mnar}. Then, we have $\Theta =\{\theta^*\} \times \Theta_{\lambda}$. That is, the ground truth model parameter $\theta^*$ can be uniquely identified via (partial) maximum likelihood learning.
\end{proposition}
\vspace{5pt}

\noindent\textbf{Missing value imputation as inference.} Given a model $p_{(\theta)}(X_o,X_u)$, the missing data imputation problem can be then formularized by the Bayesian inference problem $p_\theta(X_u|X_o) \propto p_\theta(X_u, X_o)$. If the assumptions of Proposition \ref{prop:1} are satisfied, it enables us to correctly identify the ground truth reference model parameter, $\theta^*$. Therefore, the imputed values sampled from the posterior $p_{\theta^*}(X_u|X_o)$ will be unbiased, and can be used for down stream decision making tasks.

\noindent\textbf{Remark:} Note that Proposition \ref{prop:1} can be extended to the case where model identifiability is defined by equivalence classes \cite{khemakhem2020variational, sportisse2020estimation}. 
See Appendix \ref{app:equi} for details.

\subsection{Relaxing ``correctness of parametric model'' assumption (A1)}\label{sec:relax}

In this section,  we further extend our previous results to the general MNAR cases including all different examples in Figure \ref{fig:MNAR2}. In particular, we would like to see the if the same model setting in Section \ref{sec:theory_for_h} can be applied to scenarios where $p_{\mathcal{D}}(X_o, X_u, R)$ and $p_{\theta,\lambda}(X_o, X_u, R)$ might have different parametric forms, or even different graphical representations. 

To start with, we would like to point out that the mismatch between $p_{\mathcal{D}}(X_o, X_u, R)$ and the model $p_{\theta,\lambda}(X_o, X_u, R)$ can be, to a certain extend, modeled by the \emph{mappings between spaces of parameters}. Let $\Omega \subset \mathbb{R}^I$ denote the parameter domain of our model, $p_{\theta,\lambda}(X_o, X_u, R)$. Suppose we have a mapping $\Phi: \underline{\Omega} \subset \mathbb{R}^I \mapsto \mathbb{R}^J$, such that $(\theta,\lambda) \in \underline{\Omega} \subset \Omega$ is mapped to another parameter space  $(\tau,\gamma) = \Phi(\theta,\lambda) \in \Xi \subset \mathbb{R}^J$ via $\Phi(\cdot)$. Here, $\underline{\Omega}$ is a subset of $\Omega$ on which $\Phi$ is defined. Then, the \emph{re-parameterized} $p_{\theta,\lambda}(X_o, X_u, R)$ on parameter space $\Xi$ can be rewritten as: $$ \Tilde{p}_{\tau,\gamma}(X_o, X_u, R):=p_{\Phi^{-1}(\tau,\gamma)}(X_o, X_u, R) $$
Assuming that the inverse mapping $\Phi^{-1}$ exists. Then trivially, if $p_{\theta,\lambda}(X_o,R)$ is identifiable with respect to $\theta$ and $\lambda$, then $\Tilde{p}_{\tau,\gamma}(X_o,R)$ should be also identifiable with respect to $\tau$ and $\gamma$:
\vspace{5pt}
\begin{proposition} \label{prop:2} Let $\Omega \subset \mathbb{R}^I$ be the parameter domain of the model $p_{\theta,\lambda}(X_o, X_u, R)$. Assume that the mapping $\Phi:(\theta,\lambda) \in \underline{\Omega} \subset \mathbb{R}^I \mapsto (\tau,\gamma) \in \Xi \subset \mathbb{R}^J$ is one-to-one on $\underline{\Omega}$ (equivalently, the inverse mapping $\Phi^{-1}:\Xi \mapsto \underline{\Omega}$ is injective, and $\underline{\Omega}$ is its image set). Consider the induced distribution with parameter space $\Xi$, defined as $ \Tilde{p}_{\tau,\gamma}(X_o,R):=p_{\Phi^{-1}(\tau,\gamma)}(X_o,R) $. Then, $\Tilde{p}$ is identifiable w.r.t. $(\tau,\gamma)$, if $p_{\theta,\lambda}(X_o,R)$ is identifiable w.r.t. $\theta$ and $\lambda$. 
\end{proposition} 
\vspace{5pt}

Proposition \ref{prop:2} basically shows that if two distributions $p_{\theta,\lambda}(X_o,R)$ and $ \Tilde{p}_{\tau,\gamma}(X_o,R)$ are related by a mapping $\Phi$ with nice properties, than the identifiability will translate between them. This already covers many scenarios of the data-model mismatch. For example, consider the case where ground truth data generation process satisfies the following assumption:

\noindent\textbf{Data setting D2} Suppose the ground truth $p_{\mathcal{D}}(X_o, X_u, R)$ satisfies: 
X are all generated by shared latent confounders Z (as in \textbf{D1}), and
$R$ cannot be the cause of any other variables as in \cite{mohan2013graphical, tu2019causal}.
Typical examples are given by any of the cases in Fig \ref{fig:MNAR2}(excluding (j) where $R_1$ is the cause of $R_2$). Furthermore, the ground truth data generating process is given by the parametric form $p_{\mathcal{D}}(X_o, X_u, R) = \Tilde{p}_{\tau^*,\gamma^*}(X_o, X_u, R)$, where $\Xi = \Xi_\tau \times \Xi_\gamma$ denotes its parameter space.

Then, for such ground truth data generating process, we can show that we can always find a model in the form of Equation \ref{eq:theoretical_model}, such that there exists some mapping $\Phi$, that can model their relationship:
\vspace{5pt}
\begin{lemma}\label{lem:1} Suppose the ground truth data generating process $\Tilde{p}_{\tau^*,\gamma^*}(X_o, X_u, R)$ satisfies \textbf{setting D2}. Then, there exists a model $p_{\theta,\lambda}(X_o, X_u, R)$, such that: 1), $p_{\theta,\lambda}(X_o, X_u, R)$ can be written in the form of Equation \ref{eq:theoretical_model} (i.e., \textbf{Assumption A1}; and 2), there exists a mapping $\Phi$ as described in Proposition \ref{prop:2}, such that $ \Tilde{p}_{\tau,\gamma}(X_o,R)=p_{\Phi^{-1}(\tau,\gamma)}(X_o,R)$, for all $(\tau, \gamma) \in \Xi$. 
\end{lemma}
\vspace{5pt}

\noindent\textbf{Model identification under data-model mismatch.} Since we showed the identifiability can be preserved under the parameter space mapping (Proposition \ref{prop:2}), next we will prove that if the model $p_{\theta,\lambda}(X_o, X_u, R)$ is trained on partially observed data points sampled from $\Tilde{p}_{\tau,\lambda}(X_o, X_u, R)$ that satisfies \textbf{data setting D2}, then the ML solution is still unbiased. For this purpose, inspired by Lemma \ref{lem:1},  we work with the following additional assumption:

\noindent\textbf{Model Assumption A4} Let $\Tilde{p}_{\tau^*,\gamma^*}(X_o, X_u, R)$ denote our ground truth data generating process that satisfies \textbf{data setting D2}. Then, we assume our model $p_{\theta,\lambda}(X_o, X_u, R)$ is the one that satisfies the description given by Lemma \ref{lem:1}. That is, its parametric form is given by Equation \ref{eq:theoretical_model}, and there exists a mapping $\Phi$ as described in Proposition \ref{prop:2}, such that $ \Tilde{p}_{\tau,\gamma}(X_o,R)=p_{\Phi^{-1}(\tau,\gamma)}(X_o,R)$.

%
%
Then, we have the following proposition:
\vspace{5pt}
\begin{proposition}
[Sufficient conditions for identifiability under MNAR and data-model mismatch] \label{prop:3} Let $p_{\theta,\lambda}(X_o, X_u, R)$ be a model on the observable variables $X$ and missing pattern $R$, and $p_{\mathcal{D}}(X_o, X_u, R)$ be the ground truth distribution. Assume that they satisfies \textbf{Data setting D2}, \textbf{Assumption A2, A3,} and \textbf{A4}. Let $\Theta = \arg \max_{(\theta, \lambda)\in \underline{\Omega}} \mathrm{E}_{(x_o,r) \sim p_{\mathcal{D}}(X_o,R)} \log p_{(\theta,\lambda)}(X_o = x_o,R = r)$ be the set of ML solutions of Equation \ref{eq:rubin_mnar}. Then, we have $\Theta =\{\Phi_{\tau}^{-1}(\tau^*)\} \times \Theta_{\lambda}$. Namely, the ground truth model parameter $\tau^*$ of $p_{\mathcal{D}}$ can be uniquely identified (as $\Phi(\theta^*)$) via ML learning.
\end{proposition}
\vspace{5pt}

\noindent\textbf{Remark: practical implications} Proposition \ref{prop:3} allows us to deal with the cases where the parameterization of ground truth data generating process and model distribution are related through a set of mappings, $\{\Phi_o\}$. In general, the graphical structure of $p_{\mathcal{D}}(X_o, X_u, R)$ can be any cases in Figure \ref{fig:MNAR2} excluding (j). Then, in those cases, we are still able to use a model that corresponds to Equation \ref{eq:theoretical_model} (Fig \ref{fig:MNAR2} (h)) to perform ML learning, provided that our model is flexible enough (\textbf{Assumption A4}). This greatly improves the applicability of our identifiability results, and we can build a practical algorithm based on Equation \ref{eq:theoretical_model} to handle many practical MNAR cases.

\section{GINA: A Practical Imputation Algorithm for MNAR} \label{sec:practical_model}

In the previous section, we have established the identifiability conditions for models in the form of Equation (\ref{eq:theoretical_model}). However,  in order to derive a practically useful algorithm, we still need to specify a parametric form of the model, that is both flexible and compatible with our assumptions.  In this section, by utilizing the results in Section \ref{sec:theoretical}, we propose GINA, a deep generative imputation model for MNAR data (Figure 2). GINA fulfill identifiability assumptions above, and can handle general MNAR case as discussed in section \ref{sec:relax}.   
\cz{The code is released at \url{https://github.com/microsoft/project-azua}.}

\begin{wrapfigure}[14]{r}{0.3\textwidth}
\centering
\resizebox{0.75 \linewidth}{!}{
\pgfdeclarelayer{background}
\pgfdeclarelayer{foreground}
\pgfsetlayers{background,main,foreground}

\begin{tikzpicture}

\tikzstyle{surround} = [thick,draw=black,rounded corners=1mm]
\tikzstyle{scalarnode} = [circle, draw, fill=white!11,  
    text width=1.2em, text badly centered, inner sep=2.5pt]
\tikzstyle{scalarnodenoline} = [  fill=white!11, 
    text width=1.2em, text badly centered, inner sep=2.5pt]
\tikzstyle{arrowline} = [draw,color=black, -latex]
\tikzstyle{dashedarrowcurve} = [draw,color=black, dashed, out=100,in=250, -latex]
\tikzstyle{dashedarrowline} = [draw,color=black, dashed,  -latex]

    \node [scalarnodenoline] at (1.7,0) (O)   {$\theta$};
    \node [scalarnodenoline] at (1.7,-1.5) (L)   {$\lambda$};
    \node [scalarnodenoline] at (-1.7,0) (P)   {$\phi$};
    \node [scalarnode] at (0, 0) (Z) {$\mathbf{z}_n$};
    \node [scalarnode, fill=black!30, label=below right:D,  ] at (0, -1.5) (X) {$x_nd$};
    
    \node [scalarnode, fill=black!30,  ] at (0, -3.0) (R) {$\mathbf{R}_n$};
    \node [scalarnode, fill=black!30, ] at (0, 1.0) (U) {$\mathbf{u}_n$};

    \node[surround, inner sep = .4cm,label=below right:N ] (f_N) [fit = (X)] {};
    \node[surround,fit margins={left=15pt,right=15pt,bottom=3pt,top=3pt},fit=(R) (U)]{};
    \path [arrowline] (U) to (Z);
    \path[arrowline] (Z) to (X);
    \path[arrowline]  (O) to (X);
    \path[arrowline]  (L) to (R);
    \path[arrowline]  (X) to (R);
    \path[dashedarrowline]  (P) to (Z);
    \path[dashedarrowcurve]  (-0.35,-1.5) to (-0.35,0);
    \path[dashedarrowcurve]  (-0.35,-3.0) to (-0.35,0);
\end{tikzpicture}}
\label{fig:tikz_VAE}
\caption{Graphical representations of our \name. 
} %
\end{wrapfigure}

\noindent\textbf{The parametric form of \name} We use utilize the flexibility 
of deep generative models to model the data generating process. We assume that the reference model $p_\theta(X)$ is parameterized by an identifiable VAE (see Section \ref{sec:idf_vae}) to satisfy Assumption A2. That is, $p_{\theta}(X|U) = \int_Z dZ p_\epsilon(X-f(Z)) p(Z|U)$, where $U$ is some fully observed auxiliary inputs. The decoder $p_\epsilon(X-f_\theta(Z))$ is parameterized by a neural network, $f: \mathbb{R}^H \mapsto \mathbb{R}^D$.
For convenience, we will drop the input $U$ to $p_{\theta}(X|U)$, and simply use $p_{\theta}(X)$ to denote $p_{\theta}(X|U)$. Finally, for the missing model $p_\lambda(R|X,Z)$, we use a Bernoulli likelihood model, $p_\lambda(R|X,Z):= \prod_d\pi_d(X,Z)^{R_d} (1-\pi_d(X,Z))^{1-R_d}$, where $\pi_d(X,Z)$ is parameterized by a neural network.

In Appendix \ref{app:ivae}, we show that GINA fulfill the required assumptions of Proposition \ref{prop:1} and \ref{prop:3}. Thus, we can use GINA to identify the ground truth data generating process, and perform missing value imputation under MNAR. The consistency of estimation result is also given in Appendix \ref{app:consistency}. 

\noindent\textbf{Learning and imputation} In practice, the joint likelihood in Equation \ref{eq:rubin_mnar} is intractable. Similar to the approach proposed in \cite{notmiwae}, we introduce a variational inference network, $q_\phi(Z|X_o)$, which enable us to derive a importance weighted lower bound of $\log p_{\theta,\lambda}(X_o,R)$: 
$$ \log p_{\theta,\lambda}(X_o,R) \geq \mathcal{L}_K(\theta,\lambda,\phi,X_o,R) : =  \mathbf{E}_{z^1,...,z^K, x_u^1,...,x_u^K \sim p_\theta(X_u|Z) q_\phi(Z|X_o)} \log \frac{1}{K}\sum_k w_k $$
where $w_k = \frac{p_\lambda(R|X_o ,X_u = x^k_u, Z= z^k) p_\theta(X_o,Z = z^k)}{q_\phi(Z = z^k|X_o)}$ is the importance weights. Note that we did not notate the missing pattern $R$ as additional input to $q_\phi$, as this information is already contained in $X_o$. Then, we can optimize the parameters $\theta, \lambda, \phi$ by solving the following optimization problem
$$\theta^*, \lambda^*, \phi^* = \arg \max_{\theta, \lambda, \phi} \mathbb{E}_{(x_o,r)\sim p_{\mathcal{D}} (X_o,R)} \mathcal{L}_K(\theta,\lambda,\phi,X_o = x_o, R = r)  $$
Given $\theta^*, \lambda^*, \phi^*$, we can impute missing data by solving the approximate inference problem: 
$$p_\theta(X_u|X_o) = \int_Z p_\theta(X_u|Z) p_\theta(Z|X_o)dZ \approx \int_Z p_\theta(X_u|Z) q_\phi(Z|X_o)dZ.$$
\cz{Thus, GINA can be used to predict missing data even when the data are MNAR.}

\section{Related works}
We mainly review recent works for handling MNAR data. In Appendix \ref{app:more_related}, we provide a brief review of traditional methods that deal with MCAR and MAR.

When the missing data is MNAR, a general framework is to learn a joint model on both observable variables and missing patterns \cite{little2019statistical}, in which a model of missing data is usually assumed \cite{sportisse2020imputation, ibrahim1999missing}. This approach is also widely adopted in imputation tasks. For example, in the field of recommender systems, different probabilistic models are used within such a framework \cite{hernandez2014probabilistic, marlin2009collaborative, wang2019doubly,ling2012response, liang2016modeling}. A similar approach has also been taken in the context of causal approach to imputation \cite{wang2018deconfounded, wang2019blessings, liang2016causal}. Similar to the use of the missing model, they have used an explicit model of exposure and adopted a causal view, where MNAR is treated as a confounding bias. Apart from these, inverse probability weighting methods are also used to debias the effect of MNAR \cite{schnabel2016recommendations, wang2019doubly, ma2019missing} for imputation. 

%
One issue that is often ignored by many MNAR methods is the model identifiability. Both parametric and non-parametric identifiability under MNAR has been discussed for certain cases ( \cite{miao2015identification, miao2016identifiability, miao2018identification, wang2014instrumental, tang2003analysis, sportisse2020estimation}). For example,  \cite{wang2014instrumental} proposed the instrumental variable approach to help the identification of MNAR data. \cite{miao2016identifiability} investigated the identifiability of normal and normal mixture models, and showed that identifiability for parametric models is highly non-trivial under MNAR.
\cite{miao2015identification} studied conditions for nonparametric identification using shadow variable technique. Despite the resemblance to the auxiliary variable in our approach, \cite{miao2016identifiability, miao2015identification} mainly considers the supervised learning (multivariate regression) scenario. \cite{mohan2013graphical, mohan2014graphical, shpitser2015missing} also discussed a similar topic based on a graphical and causal approach in a non-parametric setting. Although the notion of recoverability has been extensively discussed, their methods do not directly lead to practical imputation algorithms in a scalable setting. On the contrary, our work takes a different approach, in which we handle MNAR with a parametric setting, by dealing with learning and inference in latent variable models. We step aside from the computational burden with the help of recent advances in deep generative models for scalable imputation.
 
There has been a growing interest in applying deep generative models to missing data imputation. In \cite{ma2018eddi,ma2018partial, nazabal2020handling}, scalable methods for training VAEs under MAR have been proposed. 
Similar methods have also been advocated in the context of importance weighted VAEs, multiple imputation \cite{mattei2019miwae}, and heterogeneous tabular data imputation \cite{nazabal2020handling, ma2020vaem, ma2020hm}. Generative adversarial networks (GANs) have also been applied to MCAR data \cite{yoon2018gain, li2019misgan}. More recently, deep generative models under MNAR have been studied \cite{notmiwae, ghalebikesabi2021deep, gong2021variational}, where different approaches such as selection models \cite{rubin1976inference, heckman1979sample} and pattern-set mixture models \cite{little1993pattern} has been combined with partial variational inference for training VAEs. However, without additional assumptions, the model identifiability remains unclear in these approaches, and the posterior distribution of missing data conditioned on observed data might be biased.


\section{Experiments} \label{sec:exp}
\begin{figure}[t]
    \centering
    \includegraphics[width=0.9\textwidth]{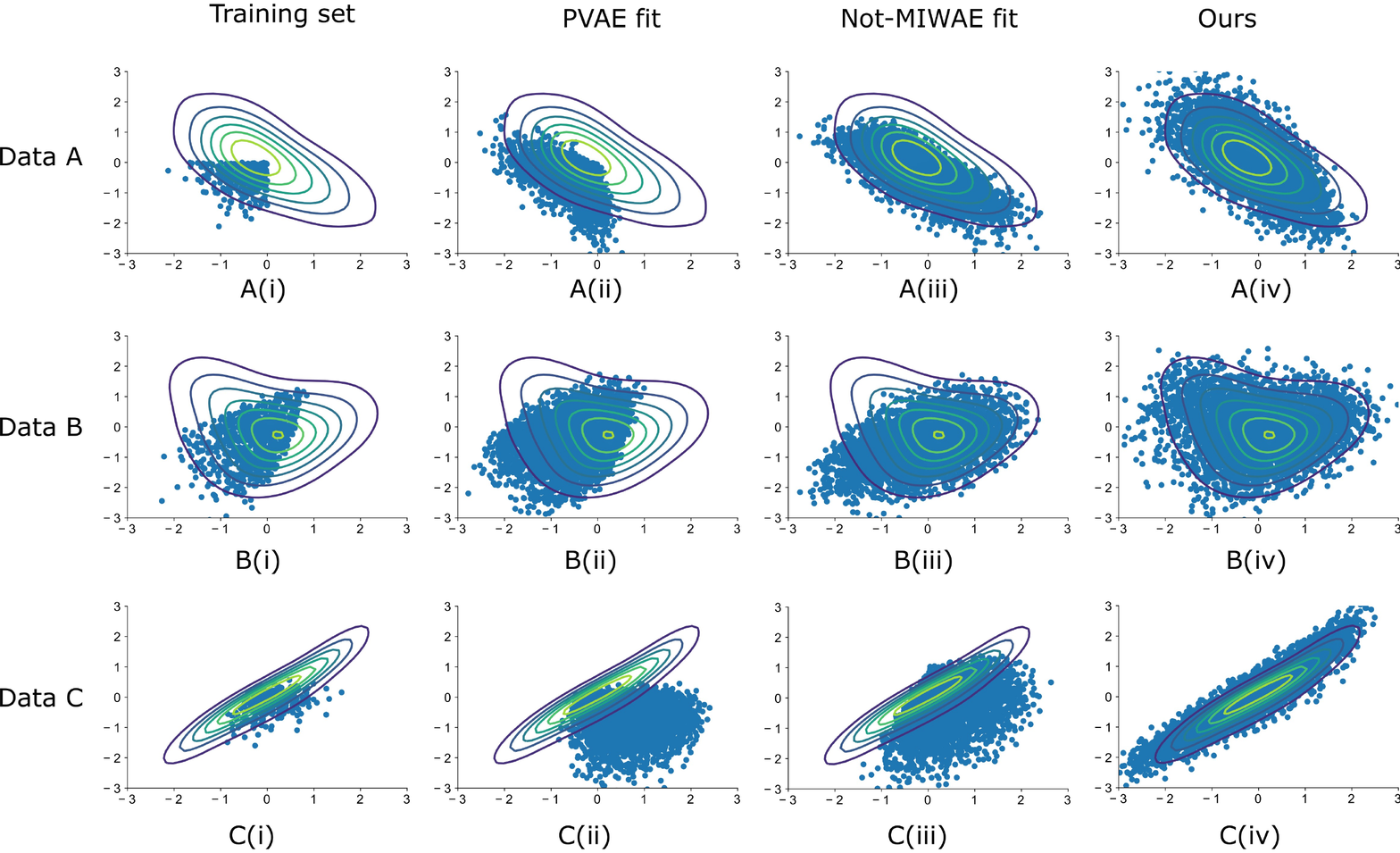}
    \caption{Visualization of generated $X_2$ and $X_3$ from synthetic experiment. 
    \textbf{Row-wise (A-C)} plots for dataset A, B, and C, respectively; \textbf{Column-wise (i-iv):} training set (only displays fully observed samples), PVAE samples, Not-MIWAE samples, and \name samples, respectively. \textbf{Contour plot}: kernel density estimate of ground truth density of complete data;
   }
    
    \label{fig:synth}
\end{figure}
We study the empirical performance of the proposed algorithm of Section 
\ref{sec:practical_model} with both synthetic data (Section \ref{sec:exp_synth}) and two real-world datasets with music recommendation (Section \ref{sec:exp_recommander}) and personalized education (Section \ref{sec:eedi}) . 
The experimental setting details can be found in Appendix \ref{app:implement}. 

\subsection{Synthetic MNAR dataset} 
\label{sec:exp_synth}

We first consider 3D synthetic MNAR datasets. We generate three synthetic datasets with nonlinear data generation process (shown in Appendix \ref{app:exp_synth}).  
 For all datasets, $X_1, X_2, X_3$ are generated via the latent variables, $Z_1, Z_2, Z_3$ ,where $X_1$ are fully observed and $X_2$ and $X_3$ are MNAR. For dataset A, we apply \emph{self-masking}(similar to Figure \ref{fig:MNAR2}(c)): $X_i$ will be missing if $X_i>0$. For datasets B and C, we apply \emph{latent-dependent self-masking}: $X_i$ will be missing, if $ g(X_i, Z_1,Z_2,Z_3) > 0$, where $g$ is a linear mapping whose coefficients are randomly chosen.

We train GINA and baseline models with partially observed data. Then, we use the trained models to generate random samples. By comparing the generated samples with the ground truth data density, we can evaluate whether $p_{\mathcal{D}}(X)$ is correctly identified. Results are visualized in Figure \ref{fig:synth}. In addition, we show the imputation results in Appendix \ref{app:additional_results}. Across three datasets, PVAE performs poorly, as it does not account for the MNAR mechanism. Not-MIWAE performs better than PVAE, as it is able to generate samples that are closer to the mode. However, it is still biased more towards the observed values. On the other hand, \name is much more aligned to ground truth, and is able to recover the ground truth from partially observed data. This experiment showed the clear advantage of our method under different MNAR situations.

\subsection{Recommender dataset imputation with random test set}
\label{sec:exp_recommander}
\begin{wraptable}[16]{l}{0.45\textwidth}
\vspace{-10pt}
\centering
\resizebox{0.4 \textwidth}{!}{
 \begin{tabular}{lcr}
 \hline
 Method & Test MSE\\
 \hline
  \hdashline
 \emph{Matrix Factorization Methods} & \\
 \hdashline
 PMF   & 1.401\\
 IPW-PMF & 1.375\\
 Deconfounded-PMF & 1.329 \\
 PMF-MNAR & 1.483 \\
 PMF-MAR & 1.480 \\
  \hdashline
  \emph{VAE-based models} & \\
  \hdashline
   PVAE & 1.259$\pm$0.003
\\
PVAE w/o IW & 1.261$\pm$0.004 \\
Not-MIWAE & 1.078$\pm$0.000\\
  \textbf{\name} & \textbf{1.052}$\pm$\textbf{0.002} \\
 \hdashline
  \emph{Others} & \\
  \hdashline
  CPTv-MNAR & 1.056 \\
  Logitvd-MNAR & 1.141 \\
  AutoRec & 1.199 \\
  Oracle-test & 1.057 \\
 \hline
 \end{tabular}
 }
  \caption{Test MSE on Yahoo! R3}
 \label{tab:yahoo}
  \begin{minipage}{0.48\textwidth}
\vspace{-10pt}
\end{minipage}
 \end{wraptable}
Next, we apply our models to recommendation systems on Yahoo! R3 dataset \cite{marlin2009collaborative, wang2018deconfounded} for user-song ratings which is designed to evaluate MNAR imputation. It contains an MNAR training set of more than 300K self-selected ratings from 15,400 users on 1,000 songs, and an MCAR test set of randomly selected ratings from 5,400 users on 10 random songs. 
We train all models 
on the MNAR training set, and evaluate 
on MCAR test set. This is repeated 10 times with different random seeds. Both the missing model for \name ($p(R|X,Z)$) and Not-MIWAE ($p(R|X)$) are parameterized by linear neural nets with Bernoulli likelihood functions. The decoders for \name, PVAE, and Not-MIWAE uses Gaussian likelihoods with the same network structure. See Appendix \ref{app:implement} for implementation details and network structures. 

We compare to the following baselines: 1), probabilistic matrix factorization (PMF) \cite{mnih2007probabilistic}; 2), inverse probability weighted PMF \cite{schnabel2016recommendations}; 3), Deconfounded PMF \cite{wang2018deconfounded}; 4), PMF with MNAR/MAR data \cite{hernandez2014probabilistic}; 5),
CPTv and Logitv models for MNAR rating \cite{marlin2009collaborative}; 6), Oracle \cite{hernandez2014probabilistic}: predicts ratings based on their marginal distribution in the test set; and 7) AutoRec \cite{sedhain2015autorec}: Autoencoders that ignores missing data. 

 Results are shown in Table \ref{tab:yahoo}. Our method (\name) gives the best performance among all methods. Also, VAE-based methods are consistently better than PMF-based methods, and MNAR-based models consistently outperform their MAR versions. 
 More importantly, among VAE-based models, our \name outperforms its non-identifiable counterpart (Not-MIWAE), and MAR counterpart (PVAE), where both models can not generate unbiased imputation. 

\subsection{Missing data imputation and active question selection on Eedi education dataset}
\label{sec:eedi}

Finally, we apply our methods to the Eedi education dataset \cite{wang2020diagnostic}, one of the largest real-world education response datasets. We consider the Eedi competition task 3 dataset, which contains over 1 million responses from 4918 students to 948 multiple-choice diagnostic questions. Each diagnostic question is a multiple-choice question. 
We consider predicting whether a student answers a question correctly or not. 
Over 70\% of the entries are missing. The dataset also contains student metadata which we use as the auxiliary variables. 
In this experiment, we randomly split the data in a 90\% train/ 10\% test/ 10\% validation ratio, and train our models on the response outcome data. 

We evaluate our model on two tasks. Firstly, we perform missing data imputation, where different methods perform imputation over the test set. As opposed to Yahoo! R3 dataset, now the test set is MNAR, thus we use the evaluation method suggested by \cite{wang2018deconfounded},  where we evaluate
the average per-question MSE For each question, over all students with non-empty response. Then, the MSEs of all questions averaged. We call this metric the debiased MSE. While regular MSE might be biased toward questions with more responses, the debiased MSE treats all questions equally, and can avoid selection bias to a certain degree. We report results for 10 repeats in the first column in Table \ref{tab:eedi}. We can see that our proposed \name achieves significantly improved results comparing to the baselines.

Secondly, we evaluate personalized education through active question selection \cite{ma2018eddi} on the test set which is task 4 from this competition dataset. The procedure is as follows: for each student in the test set, at each step, the trained generative models are used to 
decide which is the most informative missing question to collect next. This is done by maximizing the information reward as in 
\cite{ma2018eddi} (see Appendix \ref{app:eddi} for details).
Since at each step, different students might collect different questions, there isn't a simple way to debias the predictive MSE as in the imputation task. Alternatively, we evaluate each method with the help of \emph{question meta data} (difficulty level, which is a scalar).
Intuitively, when the student response to the previously collected question is correct, we expect the next diagnostic question which has higher difficulty levels, and vice versa.
Thus, we can evaluate the mean level change after correct/incorrect responses, and expect them to have significant differences. We also perform t-test between the level changes after incorrect/correct responses and report the p-value. 

\begin{table}[h]
\centering
\caption{Performance on Eedi education dataset (with standard errors)}
\scalebox{0.8}{
\begin{tabular}{p{1in}p{0.8in}p{0.8in}p{0.9in}p{0.8in}} \toprule
Method & Debiased MSE & Level change (correct) & Level change (incorrect) & p-value \\ \midrule
PVAE & 0.194$\pm$0.001 & \ 0.131$\pm$0.138 & \ -0.101$\pm$0.160 & \ 0.514\\
Not-MIWAE & 0.192$\pm$0.000 & \ 0.062$\pm$0.142 & \ -0.073$\pm$0.179 & \ 0.561\\ 
\name & \textbf{0.188}$\pm$\textbf{0.001} & \ \textbf{0.945}$\pm$\textbf{0.151} & \ \textbf{-0.353}$\pm$\textbf{0.189} & \ \textbf{1.01}$\bf{\times10^{-7}}$\\ 
\bottomrule
\end{tabular}
}
\label{tab:eedi}
\end{table}
We can see in Table \ref{tab:eedi}, \name is the only method that reports a significant p-value (<0.05) between the level changes of next collected questions after incorrect/correct responses which are desired. This further indicates that our proposed GINA predicts the unobserved answer with the desired behavior. 
\section{Conclusion}
In this paper, we provide a analysis of identifiability for generative models under MNAR, and studies sufficient conditions of identifiability under different scenarios. We provide sufficient conditions under which the model parameters can be uniquely identified, via joint maximum likelihood learning on $X_o$ and $R$. Therefore, the learned model can be used to perform unbiased missing data imputation. We proposed a practical algorithm based on VAEs, which enables us to apply flexible generative models that is able to handle missing data in a principled way. The main limitation of our proposed practical algorithm is the need for auxiliary variables $U$, which is inherited from identifiable VAE models \cite{khemakhem2020variational}. In practice, they may not be always available. For future work, we will investigate how to address such limitation, and how to extend to more complicated scenarios.

\newpage
\section*{Acknowledgements and Disclosure of Funding} 
We thank Martin Kukla, Angus Lamb, Yingzhen Li, Dawen Liang, Chang Liu,  Ruibo Tu, Yordan Zaykov (in alphabetical order) for helpful discussions and implementation support. 
\bibliographystyle{plain}
\bibliography{REF}
\appendix
\newpage
\section{Traditional methods for handling missing data} \label{app:more_related}
Methods for handling missing data has been extensively studied in the past few decades. Those methods can be roughly classified into two categories: complete case analysis (CCA) based, and imputation based methods. CCA based methods, such as listwise deletion \cite{allison2001missing} and pairwise deletion \cite{marsh1998pairwise} focuses on deleting data instances that contains missing entries, and keeping those that are complete. Listwise/pairwise deletion methods are known to be unbiased under MCAR, and will be biased under MAR/MNAR. On the contrary, imputation based methods tries to replace missing values by imputed/predicted values. One popular imputation technique is called single imputation, where only produce one single set of imputed values for each data instance. Standard techniques of single imputation include mean/zero imputation, regression-based imputation \cite{allison2001missing}, no- parametric methods \cite{keerin2012cluster, stekhoven2012missforest}. As opposed to single imputation, the multiple imputation (MI) methods such as MICE \cite{white2011multiple}, was first proposed by Rubin \cite{rubin1977formalizing, rubin1988overview, horton2001multiple, murray2018multiple} is essentially a simulation-based methods that returns multiple imputation values for subsequent statistical analysis. Unlike single imputation, the standard errors of estimated parameters produced with MI is known to be unbiased \cite{rubin2004multiple}. Apart from MI, there exists other methods such as full information maximum likelihood \cite{arbuckle1996full, enders2001relative} and inverse probability weighting \cite{robins1994estimation, horvitz1952generalization}, which can be directly applied to MAR without introducing additional bias. However, these methods assumes a MAR missing data mechanism, and cannot be directly applied to MNAR without introducing bias.

\section{Implementation details} \label{app:implement}

We first introduce the general settings of \name and other baselines. Our model (\name) is based on the practical algorithm in Section \ref{sec:practical_model}. By default, we will set the auxiliary variable $U$ to be some fully observed meta feature (if there's any) or the missing mask pattern (if the dataset does not have a fully observed meta feature). The most important baselines are as follows: 1), Partial VAE (PVAE) \cite{ma2018eddi}: a VAE model with slightly modified ELBO objective, specifically designed for MAR data; and 2), Not-MIWAE \cite{notmiwae}, a VAE model for MNAR data trained by jointly maximizing the likelihood on both the partially observed data and the missing pattern. As opposed to our model, the latent priors $p(Z)$ for both PVAE and Not-MIWAE are parameterized by a standard normal distribution, hence no auxiliary variables are used. Also, note that the graphical model of Not-MIWAE is described by Fig \ref{fig:MNAR2} (d), and does not handle the scenarios where the ground truth data distribution follows other graphs like Fig \ref{fig:MNAR2} (g). Finally, the inference model $q(Z|X)$ for the underlying VAEs is set to be diagonal Gaussian distributions whose mean and variance are parameterized by neural nets as in standard VAEs \cite{kingma2013auto} (with missing values replaced by zeros\cite{nazabal2020handling, notmiwae, mattei2019miwae}), or a permutation invariant set function proposed in \cite{ma2018eddi}. See Appendix \ref{app:implement} for more implementation details for each tasks.

\subsection{Synthetic dataset implementation details} \label{app:exp_synth}

\paragraph{Data generation} The ground truth data generating process is given by 
$
    Z_1, Z_2, Z_3 \sim \mathcal{N}(0,1), 
    X_1 = h_{w}(Z_1,Z_2,Z_3) + \epsilon_1, 
    X_2 = f_{\theta_1}(X_1,Z_1,Z_2,Z_3) + \epsilon_2,
    X_3 = f_{\theta_2}(X_1,X_2, Z_1,Z_2,Z_3) + \epsilon_3 $
where $h_w$ is a linear mapping with coefficients $w$, $f$ is some non-linear mapping whose functional form is given by Appendix \ref{app:implement}, $\theta_1$ \& $\theta_2$ are two different sets of parameters for $f$, and $\epsilon_1$, $\epsilon_2$, $\epsilon_3$ are observational noise variables with mean 0, variance 0.01. We randomly sample three different sets of parameters, and generate the corresponding datasets (Figure \ref{fig:synth}), namely dataset A, B, and C. Each dataset consists of 2000 samples. Then, we apply different missing mechanisms for each dataset. For all datasets, we assume that $X_1$ are fully observed and $X_2$ and $X_3$ are MNAR. , and missing mechanism will be only applied to $X_2$ and $X_3$. Finally, all observable variables are standardized.

\paragraph{Remark} Note that in Dataset A, the ground truth missing mechanism does not depend on the latent variable model. Therefore, in this case, the not-MIWAE model does not have model-misspecification problem, hence any less satisfying performance is due to non-identifiability.

\paragraph{Network structure and training} 
We use 5 dimensional latent space with fully factorized standard normal priors. The decoder part $p_\theta(X|Z)$ uses a 5-10-$D$ structure, where $D = 3$ in our case.  For inference net, we use a zero imputing \cite{ma2018eddi} with structure 2$D$-10-10-5, that maps the concatenation of observed data (with missing data filled with zero) and mask variable $R$ into distributional parameters of the latent space. For the factorized prior $p(Z|U)$ of the i-VAE component of \name, we used a linear network with one auxiliary input (which is set to be fully observed dimension, $X_1$). The missing model $p_\lambda(R|X)$ for \name and i-NotMIWAE is a single layer neural network with 10 hidden units. All neural networks use Tanh activations (except for output layer, where no activation function is used). All baselines uses importance weighted VAE objective with 5 importance samples. The observational noise for continuous variables are fixed to $\log \sigma = -2$. All methods are trained with Adam optimizer with batchsize with 100, and learning rate 0.001 for 20k epochs. 

\subsection{Yahoo!R3 experiment implementation details}

Before training, all user ratings are scaled to be between 0 and 1 (such scaling will be reverted during evaluation). For all baselines,  we use Gaussian likelihood with variance of 0.02. We use 20 dimensional latent space, and the decoder $p_\theta(X|Z)$ uses a 20-10-$D$ structure. We use Tanh activation function for the decoder (except for output layer, where no activation function is used). For inference net, we uses the point net structure proposed in \cite{ma2018eddi}, we use 20 dimensional feature mapping $h$ parameterized by a single layer neural network and 20 dimensional ID vectors for each variable. The symmetric operator is set to be the summation operator. The missing model $p_\lambda(R=1|X)$ for \name and i-NotMIWAE is parameterized by linear neural network. All methods are trained with 400 epochs with batchsize 100.

\subsection{Eedi dataset experiment implementation details}

Since Eedi dataset is a binary matrix with 1/0 indicating that the student response is correct/incorrect, we use Bernoulli likelihood for decoder $p_\theta(X|Z)$. For We use 50 dimensional latent space, and the decoder $p_\theta(X|Z)$ uses a 50-20-50-$D$ structure. Such structure is chosen using the validation set using grid search. We use ReLU activation function for the decoder (except for output layer, where no activation function is used). For inference net, we uses the point net structure that were used in Yahoo!R3 dataset. Here, the difference is that we we use 50 dimensional feature mapping $h$ parameterized by a single layer neural network and 10 dimensional ID vectors for each variable.  All methods are trained with 1k epochs with batchsize 100. A trick that we used for both not-MIWAE and \name to improve the imputation performance, is to turn down the weight of the likelihood term for $p_\lambda(R|X)$, by multiplying a factor of $\beta = 0.5$. This is due to that majority of the student response matrix is missing, the $p_\lambda(R|X)$ will most likely dominate the training, hence the learning algorithm will prefer more about learning the models that explains the missing mechanism better, over the models that explains the observable variables $X$ better.

\section{Proof for Proposition \ref{prop:1}} \label{app:proof1}

\paragraph{Proof}: 
First, we show that $p_{\theta,\lambda}(X_{O'_l},R)$ is \emph{partially identifiable} (i.e., identifiable on subset of parameters) on $\{\theta_d\}_{d\in O'_l}$ for $\forall O'_l \in \bar{\mathcal{A}}_\mathcal{I}$. We prove this by contradiction. Suppose there exists two different set of parameters $(\theta^1,\lambda^1)$ $(\theta^2,\lambda^2)$, such that there exits at least one index $c\in O'_l$ for some $l$, such that $\theta_c^1 \neq \theta_c^2$, and $p_{(\theta^1,\lambda^1)}(X_{O'_l},R) = p_{(\theta^2,\lambda^2)}(X_{O'_l},R)$. That is, $p(X_{O'_l},R)$ is not identifiable on $\{\theta_d\}_{d\in O'_l}$. 

According to \textbf{Assumption A3}, there exists $O_s \in {\mathcal{A}}_\mathcal{I}$, such that $c\in O_s \subset O'_l$. Then, consider the marginal $$p_{\theta}(X_{O_s}) = \int_{Z,R,X_{\setminus O_s}} dZ  \prod_{d\in O_s} p_{\theta_d}(X_d|Z) p_{\lambda}(R|X,Z) p(Z) = p_{\theta_{d \in O_s}}(X_{O_s}) $$. Since $p_{(\theta^1,\lambda^1)}(X_{O'_l},R) = p_{(\theta^2,\lambda^2)}(X_{O'_l},R)$, we have $p_{(\theta_{O_s}^1)}(X_{O_s}) = p_{(\theta_{O_s}^2)}(X_{O_s})$ (the joint uniquely determines marginals).
However, this contradicts with our \textbf{Assumption A2} that $p_{\theta_{O_s}}(X_{O_s})$ is identifiable: this identifiability assumption implies that we should have  $p_{(\theta_{O_s}^1)}(X_{O_s}) \neq p_{(\theta_{O_s}^2)}(X_{O_s})$. Therefore, by contradiction, we have $p(X_{O'_l},R)$ is partially identifiable on $\{\theta_d\}_{d\in {O'_l}}$ for $\forall O'_l \subset \bar{\mathcal{A}}_\mathcal{I}$.

Then, we proceed to prove that the ground truth parameter $\theta^*$ can be uniquely identified via ML learning. Based on our \textbf{Assumption A1}, 
%
%
upon optimal ML solution, 
$$\Theta_{ML} = \arg \max_{(\theta, \lambda)\in \Omega} \mathrm{E}_{(x_o,r) \sim p_{\mathcal{D}}(X_o,R)} \log p_{(\theta,\lambda)}(X_o = x_o,R = r)$$, we have the following identity:
$$p_{(\theta_{ML},\lambda_{ML})}(X_o,R) = p_{(\theta^*,\lambda^*)}(X_o,R) $$ holds for all $(\theta_{ML},\lambda_{ML}) \in \Theta_{ML}$, and all $\forall O \subset \mathcal{I}$ that satisfies $p(X_o, X_u, R_o=1, R_u = 0)>0$.

Note also that:

$$p_{(\theta_{ML},\lambda_{ML})}(X_o,R) = \int_{Z,X_{\mathcal{I} \setminus O}} dZ \prod_{d} p_{\theta^{ML}_d}(X_d|Z) p_{\lambda^{ML}}(R|X) p(Z) $$

, which depends on both $\theta_o$ and $\lambda$. Since we have already shown that $p_{(\theta,\lambda)}(X_{O'_l},R)$ are partially identifiable on $\{\theta_d\}_{d\in {O'_l}}$ for $\forall {O'_l} \subset \bar{\mathcal{A}}_{\mathcal{I}}$ and according to \textbf{Assumption A3}, $p_{\mathcal{D}}(X_o, X_u, R_{O'_l}=1, R_{\mathcal{I}\setminus O'_l} = 0)>0$. Therefore, upon optimal solution 

, we have that 
 $$ \{\theta_d = \theta_{d}^* \}_{d \in {O'_l}}$$
 
 holds for all $\forall {O'_l} \subset \bar{\mathcal{A}}_{\mathcal{I}}$. Since we have assumed that $\bigcup_{{O'_l}\in \bar{\mathcal{A}}_{\mathcal{I}}} X_{O'_l} = \mathcal{I}$ in \textbf{Assumption 3} (i.e.,$\bar{\mathcal{A}}_{\mathcal{I}}$ is a cover of $\mathcal{I}$ ), this guarantees that 

$$\theta^{ML}_d = \theta_{d}^* $$ for all $d$. In other words, we are able to uniquely identify $\theta^*$ from observed data, therefore $$\Theta =\{\theta^*\} \times \Theta_{\lambda}$$.
\qed

\paragraph{Remark (examples).} To better illustrated the implication of the proposition, we provide an example that satisfies the assumptions of Proposition \ref{prop:1}. One example is the self-masking one-way ANOVA \cite{little2019statistical}, which contains $I$ observable variables, $X = \{X_1,...,X_I\}$, generated according to
\begin{equation*}
    X_i|Z_i \sim \mathcal{N}(X_i;Z_i,\sigma^2), \quad i = 1,...,I,
\end{equation*}
where $\sigma^2$ is some known observational noise variance, $Z = \{Z_1,...,Z_I\}$ are latent variables generated by 
\begin{equation*}
    Z_i \sim \mathcal{N}(Z_i;\mu_i,\varsigma^2).
\end{equation*}
Hence, the learnable parameter $\theta$ is given by $\theta = \{\mu_1,...,\mu_I, \varsigma^2\}$. Then, assume that each observable variable $X_i$ is missing MNAR according to the following mechanism:
\begin{equation*}
    p(R_i=1|X, Z) = \text{Sigmoid}(\lambda_1 Z_i + \lambda_0),
\end{equation*}
where $\lambda = \{\lambda_0, \lambda_1\}$ is the set of learnable parameters for missing mechanism. Now, we verify that $p_{\theta,\lambda}(X,Z)$ satisfies the assumptions of proposition \ref{th:1}:
\begin{itemize}
    \item \textbf{Assumption A2}. By taking $\mathcal{A}_\mathcal{I} = \{O_s\}_{1\leq s \leq I}$, where $O_s = \{s\}$, we have $p_\theta(X_s) = \mathcal{N}(X_s; \mu_s, \sigma^2 + \varsigma^2)$, which is identifiable over the subset of parameters $\theta_s =\{ \mu_s, \varsigma^2  \}$. Note that the partition $\mathcal{A}_\mathcal{I} = \{O_s\}_{1\leq s \leq I}$ is not unique; in fact, since $p_\theta(X_{O_s}) = \prod_{i \in O_s} \mathcal{N}(X_i; \mu_i, \sigma^2 + \varsigma^2)$ is identifiable on $\{\mu_s|s\in O_s\} \bigcup \{\varsigma^2\}$ for all non-empty $O_s \subset \mathcal{I}$, any partition $\mathcal{A}_\mathcal{I} = \{O_s\}_{1\leq s \leq I}$ will satisfy \textbf{Assumption A2}.
    \item \textbf{Assumption A3}. Since $p_{\theta}(X)$ and $p_\lambda(R|X, Z)$ are strictly positive for all possible settings of $X$, $R$, $\theta$ and $\lambda$, \textbf{Assumption A3} is trivially satisfied.
\end{itemize}

\section{Proof of Proposition \ref{prop:2}} \label{app:proof2}

\paragraph{Proof} Let $(\tau_1,\gamma_1)$ and $(\tau_2,\gamma_2)$ be two different parameters in $\Xi$. Then, we have
\begin{align*}
    &\Tilde{p}_{\tau_1,\gamma_1}(X_o,R) \\ 
    =& p_{\Phi^{-1}(\tau_1,\gamma_1)}(X_o,R) \\
    \neq & p_{\Phi^{-1}(\tau_2,\gamma_2)}(X_o,R) \\
    =& \Tilde{p}_{\tau_2,\gamma_2}(X_o,R) 
\end{align*}
where the third line is due to the fact that $\Phi^{-1}$ is injective and $p_{\theta,\lambda}(X_o,R)$ is identifiable with respect to $\theta$ and $\lambda$.
\qed

\section{Relaxing Assumption A1}\label{app:relax}

\subsection{Proof of Lemma \ref{lem:1}}

\begin{customlemma}{1} Suppose the ground truth data generating process $\Tilde{p}_{\tau^*,\gamma^*}(X_o, X_u, R)$ satisfies \textbf{setting D2}. Then, there exists a model $p_{\theta,\lambda}(X_o, X_u, R)$, such that: 1), $p_{\theta,\lambda}(X_o, X_u, R)$ can be written in the form of Equation \ref{eq:theoretical_model} (i.e., \textbf{Assumption A1}; and 2), there exists a mapping $\Phi$ as described in Proposition \ref{prop:2}, such that $ \Tilde{p}_{\tau,\gamma}(X_o,R)=p_{\Phi^{-1}(\tau,\gamma)}(X_o,R)$, for all $(\tau, \gamma) \in \Xi$. Additionally, such $\Phi$ is decoupled, i.e., $\Phi(\theta,
\lambda) = (\Phi_{\theta}(\theta),\Phi_{\lambda}(\lambda))$.
\end{customlemma}

\paragraph{Proof:} \footnote{We mainly consider the case where all variables are continuous. Discrete variables will complicate the discussion, but will not change the conclusion.}

\paragraph{Case 1 (connections among $X$):} Suppose the ground truth data generating process $p_{\mathcal{D}}(X,R) = \Tilde{p}_{\tau^*,\gamma^*}(X_o, X_u, R)$ is given by Figure \ref{fig:MNAR2} (i). That is, $p_{\mathcal{D}}(X,R) = \Tilde{p}_{\gamma}(X|R) \int_Z \prod_i \Tilde{p}_{\tau_i}(X_i|Z, pa(X_i) \bigcap X) p(Z) dZ$. Without loss of generality, assume that probabilistic distributions $\Tilde{p}_{\tau_i}(X_i|Z, pa(X_i) \bigcap X)$ takes the form as
%
%
$\Tilde{p}_{\tau_i}(X_i|Z, pa(X_i) \bigcap X) = \int_{\epsilon_i} \delta(X_i - f_i^{\lambda_i}(\epsilon_i, pa(X_i)\bigcap X, Z)) p(\epsilon_i) d\epsilon_i$. Therefore, we have 


\begin{align*}
    &\Tilde{p}_{\tau}(X) \\
    =& \int_Z \prod_i \Tilde{p}_{\tau_i}(X_i|Z, pa(X_i) \bigcap X) p(Z) dZ \\
    =& \int_z \left[\prod_{\{i| N(X_i) \bigcap X \neq \emptyset\}} \int_{ \epsilon_i } d\epsilon_i \delta(X_i - f_i^{\lambda_i}(\epsilon_i, pa(X_i)\bigcap X, Z)) p(\epsilon_i)\right] \\
    &\left[\prod_{\{j|N(X_j)\bigcap X = \emptyset \}}p(X_j|Z) \right] p(Z) dZ \\
    =& \int_{z, \{i|N(X_i)\bigcap X \neq \emptyset \}} \left[\prod_{\{i| N(X_i) \bigcap X \neq \emptyset\}} \delta(X_i - f_i^{\lambda_i}(\epsilon_i, pa(X_i)\bigcap X, Z)) p(\epsilon_i)\right] \\
    &\left[\prod_{\{j|N(X_j)\bigcap X = \emptyset \}}p(X_j|Z) \right]   p(Z) dZ \\
\end{align*}


Apparently, there exists a set of function $\{g_i(\cdot)|N(X_i)\bigcap X \neq \emptyset \}$, such that:
\begin{align*}
& \int_{z, \{i|N(X_i)\bigcap X \neq \emptyset \}} \left[\prod_{\{i| N(X_i) \bigcap X \neq \emptyset\}} \delta(X_i - f_i^{\lambda_i}(\epsilon_i, pa(X_i)\bigcap X, Z)) p(\epsilon_i)\right] \\
&\left[\prod_{\{j|N(X_j)\bigcap X = \emptyset \}}p(X_j|Z) \right]   p(Z) dZ \\
=& \int_{z, \{i|N(X_i)\bigcap X \neq \emptyset \}} \left[\prod_{\{i| N(X_i) \bigcap X \neq \emptyset\}} \delta(X_i - g_i(\epsilon_i, anc_\epsilon(i), Z)) p(\epsilon_i)\right] \\
&\left[\prod_{\{j|N(X_j)\bigcap X = \emptyset \}}p(X_j|Z) \right]   p(Z) dZ 
\end{align*}
Where $anc_\epsilon(i)$ is the shorthand for 

$$\{ \epsilon_k | X_k\in anc{X_i}\bigcap Z, 1\leq k \leq D  \}$$

Note that, the graphical model of the new parameterization, 
\begin{align*}
    p(X) = & \int_{z, \{i|N(X_i)\bigcap X \neq \emptyset \}} \left[\prod_{\{i| N(X_i) \bigcap X \neq \emptyset\}} \delta(X_i - g_i(\epsilon_i, anc_\epsilon(i), Z)) p(\epsilon_i)\right]\\
    &\left[\prod_{\{j|N(X_j)\bigcap X = \emptyset \}}p(X_j|Z) \right]   p(Z) dZ 
\end{align*}
has a new aggregated latent space, $\{Z,  \{\epsilon_i|1\leq i\leq D \}\}$. That is, for each $X_i$ that has non empty neighbour in $X$, a new latent variable will be created. With this new latent space, the connections among $X$ can be decoupled, and the new graphical structure of $p(X,R)$ corresponds to Figure \ref{fig:MNAR2} (h). 

The mapping $\Phi$ that connects $\Tilde{p}_{\tau_i}(X,R)$ and $p(X,R)$ can now be defined as identity mapping, since no new parameters are introduced/removed when reparameterizing $\Tilde{p}_{\tau_i}(X,R)$ into $p(X,R)$. Hence, the two requirements of Lemma \ref{lem:1} are fulfilled.

\paragraph{Case 2(subgraph):} Next, consider the case that the ground truth data generating process $p_{\mathcal{D}}(X,R) = \Tilde{p}_{\tau^*,\gamma^*}(X_o, X_u, R)$ is given by one of the Figure \ref{fig:MNAR2} (a)-(g). That is, it is a subgraph of Figure \ref{fig:MNAR2} (h). Without loss of generality, assume that $\Tilde{p}_{\gamma_i}(R_i=1|pa(R_i)) = \mathrm{logit^{-1}( f_{\gamma_i}(pa(R_i)) )}$, and $pa(R_i) \subsetneq \{X,Z\}$; in other words, certain connections from $\{X,Z\}$ to $R_i$ is missing. Consider the model distribution parameterized by ${p}(R_i=1|X,Z) = \mathrm{logit^{-1}( f_{\gamma_i}(pa(R_i)) + g_{\theta_i}( \{X,Z\} \setminus  pa(R_i)) )}$, satisfying $g_{\theta_i=0}(\cdot) \equiv 0$. Therefore, the mapping $\Phi^{-1}$ is given as $\Phi^{-1}(\gamma_i) := (\gamma_i,\theta_i=0)$. Apparently, $\Phi^{-1}$ is injective, hence satisfying the requirement of Proposition \ref{prop:2}.

\qed

\subsection{Proof for Proposition \ref{prop:3}} \label{app:proof3}

\begin{customproposition}{3}
[Sufficient conditions for identifiability under MNAR and data-model mismatch]  Let $p_{\theta,\lambda}(X_o, X_u, R)$ be a model on the observable variables $X$ and missing pattern $R$, and $p_{\mathcal{D}}(X_o, X_u, R)$ be the ground truth distribution. Assume that they satisfies \textbf{Data setting D2}, \textbf{Assumption A2, A3,} and \textbf{A4}. Let $\Theta = \arg \max_{(\theta, \lambda)\in {\Omega}} \mathrm{E}_{(x_o,r) \sim p_{\mathcal{D}}(X_o,R)} \log p_{(\theta,\lambda)}(X_o = x_o,R = r)$ be the set of ML solutions of Equation \ref{eq:rubin_mnar}. Then, we have $\Theta =\{\Phi_{\tau}^{-1}(\tau^*)\} \times \Theta_{\lambda}$. Namely, the ground truth model parameter $\tau^*$ of $p_{\mathcal{D}}$ can be uniquely identified (as $\Phi(\theta^*)$) via ML learning.
\end{customproposition}

\paragraph{Proof}: 
First, it s not hard to show that $p_{\theta,\lambda}(X_{O'_l},R)$ is partially identifiable on $\{\theta_d\}_{d\in O'_l}$ for $\forall O'_l \in \bar{\mathcal{A}}_\mathcal{I}$. This has been shown in the proof of Proposition \ref{prop:1}, and we will not repeat this proof again. 




Next, given data setting \textbf{D2} and \textbf{Assumption A4}, define 
$$\Theta_{ML} = \arg \max_{(\theta, \lambda)\in {\Omega}} \mathrm{E}_{(x_o,r) \sim p_{\mathcal{D}}(X_o,R)} \log p_{(\theta,\lambda)}(X_o = x_o,R = r)$$, then we have:
$$p_{(\theta_{ML},\lambda_{ML})}(X_o,R) = p_{\Phi^{-1}(\tau^*,\gamma^*)}(X_o,R) $$ holds for all $(\theta_{ML},\lambda_{ML}) \in \Theta_{ML}$, and all $\forall O \subset \mathcal{I}$ that satisfies $p(X_o, X_u, R_o=1, R_u = 0)>0$.


Since $p_{(\theta,\lambda)}(X_{O'_l},R)$ are partially identifiable on $\{\theta_d\}_{d\in {O'_l}}$ for $\forall {O'_l} \subset \bar{\mathcal{A}}_{\mathcal{I}}$ and according to \textbf{Assumption A3}, $p_{\mathcal{D}}(X_o, X_u, R_{O'_l}=1, R_{\mathcal{I}\setminus O'_l} = 0)>0$. Therefore,
 $$ \{\theta_d = \Phi^{-1}_{\theta}(\tau^*,\gamma^*)_{d} \}_{d \in {O'_l}}$$
must be true for all $\forall {O'_l} \subset \bar{\mathcal{A}}_{\mathcal{I}}$, where $\Phi^{-1}_{\theta}(\tau^*,\gamma^*)$ denotes the components of $\Phi^{-1}(\tau^*,\gamma^*)$ that corresponds to the entries of $\theta$. Since we have assumed that $\bigcup_{{O'_l}\in \bar{\mathcal{A}}_{\mathcal{I}}} X_{O'_l} = \mathcal{I}$ in \textbf{Assumption 3} (i.e.,$\bar{\mathcal{A}}_{\mathcal{I}}$ is a cover of $\mathcal{I}$ ), this guarantees that 

$$\theta^{ML}_d = \Phi^{-1}_{\theta}(\tau^*,\gamma^*)_d $$ for all $d$. In other words, we are able to uniquely identify $\theta^*$ from observed data, therefore $$\Theta =\{\Phi^{-1}_{\theta}(\tau^*,\gamma^*)\} \times \Theta_{\lambda}$$.

Finally, according to \textbf{Assumption 4} and the proof of Lemma \ref{lem:1}, $\Phi$ is decoupled as $(\Phi_{\theta}(\theta),\Phi_{\lambda}(\lambda))$. Therefore, we can write
$\Theta =\{\Phi^{-1}(\tau^*)\} \times \Theta_{\lambda}$. That is, the ground truth model parameter $\tau^*$ of $p_{\mathcal{D}}$ can be uniquely identified (as $\Phi(\theta^*)$).

\qed

\section{Identifiability based on equivalence classes} \label{app:equi}

In this section, we introduce the notion of identifiability based on equivalence classes. Let $\sim$ be a equivalence relation on a parameter space $\Omega$. That is, it satisfies reflexivity ($\theta_1 \sim \theta_1$), symmetry ($\theta_1 \sim \theta_2$ if and only if $\theta_2 \sim \theta_1$), and transitivity (if $\theta_1 \sim \theta_2$ and $\theta_2 \sim \theta_3$, then $\theta_1 \sim \theta_3$). Then, a equivalence class of $\theta_1 \in \Omega$ is defined as $\{\theta|\theta\in \Omega, \theta \sim \theta\}$. We denote this by $[\theta_1]$. Then, we are able to give the definition of model identifiability based on equivalence classes:

\begin{definition}[Model identifiability based on equivalence relation]\label{def:idf3} Assume $p_\theta(X)$ is a distribution of some random variable $X$, $\theta$ is its parameter that takes values in some parameter space $\Omega_\theta$, and $sim$ some equivalence relation on $\Omega$ Then, if $p_\theta(X)$ satisfies $
p_{\theta_1}(X) = p_{\theta_2}(X) \Longleftrightarrow \theta_1 \sim \theta_2 \Longleftrightarrow [\theta_1] = [\theta_2], \forall \theta_1,\theta_2 \in \Omega_\theta$, 
we say that $p_\theta$ is \emph{$\sim$ identifiable} w.r.t. $\theta$ on $\Omega_\theta$.
\end{definition}

Apparently, definition \ref{def:idf} is a special case of definition \ref{def:idf3}, where $\sim$ is given by the equality operator, $=$. When the discussion is based on the identifiability under equivalence relation, then it is obvious that all the arguments of Proposition \ref{prop:1}, \ref{prop:2}, and \ref{prop:3} still holds. Also, the statement of the results needs to adjusted accordingly.  For example, in Proposition \ref{prop:1}, instead of ``the ground truth model parameter $\theta^*$ can be uniquely identified", we now have ``the ground truth model parameter $\theta^*$ can be uniquely identified \emph{up to a equivalence relation, $\sim$}".

\section{Subset identifiability (A2) for identifiable VAEs} \label{app:ivae}

The GINA model needs satisfy the requirement on model of Proposition \ref{prop:1} or \ref{prop:3}, if we wish to use it to fit to the partially observed data and then perform (unbiased) missing data imputation. In order to show that the identifiability result of Proposition \ref{prop:1}/\ref{prop:3} can be applied to GINA, the key assumption that we need to verify is the local identifiability (\textbf{Assumption A2}).

To begin with, in \cite{khemakhem2020variational}, the following theorem on VAE identifiability has been proved:

\begin{theorem}\label{th:1} Assume we sample data from the model given by $p(X,Z|U) = p_{\epsilon}(X-f(Z))p_{T,\zeta}(Z|U)$, where $f$ is a multivariate function $f: \mathbb{R}^H \mapsto \mathbb{R}^D$. $p_{T,\zeta}(Z|U)$ is parameterized by exponential family of the form $p_{T,\zeta}(Z|U) \propto \prod_{i=1^M} Q(Z_i)\exp[ \sum_{j=1^K}T_{i,j}(Z_i)\zeta_{i,j}(U) ]$, where $Q(Z_i)$ is some base measure, $M$ is the dimensionality of the latent variable $Z$, $\mathbf{T}_i(U) = ( T_{i,1}, ..., T_{i,K} )$ are the sufficient statistics, and $\bm{\zeta}_i(U) = ( \zeta_{i,1}, ..., \zeta_{i,K} )$ are the corresponding parameters, depending on $U$. Assume the following holds:
\begin{enumerate}
    \item The set $\{X\in\mathcal{X}| \phi_\epsilon(x) = 0\}$ has zero measure, where $\phi$ is the characteristic function of $p_{\epsilon}$;
    \item The multivariate function $f$ is injective;
    \item $T_{i,j}$ are differentiable a.e., and $(T_{i,j})_{1\leq j\leq k}$ are linearly independent on any subset of $\mathcal{X}$ of measure greater than zero;
    \item There exists $nk+1$ distinct points $U^0,...,U^{nk}$, such that the matrix $L = (\bm{\zeta}(U^1-U^0),...,\bm{\zeta}(U^{nk}-U^0))$ of size $nk$ by $nk$ is invertible.
\end{enumerate}
Then, the parameters $(f,T,\zeta)$ are $\sim_{A}$-identifiable, where $\sim_{A}$ is the equivalence class defined as (see also Appendix \ref{app:equi}):
$$ (f,T,\zeta) \sim (\Tilde{f},\Tilde{T},\Tilde{\zeta}) \iff \exists \mathbf{A},\mathbf{c} | \mathbf{T}(f^{-1}(X)) = \mathbf{A}\mathbf{T}(\Tilde{f}^{-1}(X))+\mathbf{c} $$. Here, $\mathbf{A}$ is a $nk$ by $nk$ matrix, and $\mathbf{c}$ is a vector.
\end{theorem}

Note that under additional mild assumptions, the $\mathbf{A}$ in the $
\sim_A$ equivalence relation can be further reduced to a permutation matrix. That is, the model parameters can be identified, such that the latent variables differs up to a permutation. This is inconsequential in many
applications. We refer to \cite{khemakhem2020variational} for more discussions on permutation equivalence.

So far, Theorem \ref{th:1} only discussed the identifiability of $p(X)$ on the \emph{full} variables, $X = X_o \bigcup X_u$. However, in \textbf{Assumption A2}, we need the reference model to be (partially) identifiable on a partition $O_s \in \mathcal{A}_\mathcal{I}$, $p_\theta(X_{o_s})$. Naturally, we need additional assumptions on the the injective function $f$, as stated below:
\paragraph{Assumption A5} There exists an integer $D_o$, such that $f_O: \mathbb{R}^H\mapsto \mathbb{R}^{|O|}$ is injective for all $O$ that $|O|\geq D_0$. Here, $f_O$ is the entries from the output of $f$, that corresponds to the index set $O$. 



\paragraph{Remark} Note that, under assumption $\textbf{A5}$, the Assumption \textbf{A3} in Section \ref{sec:theoretical} becomes more intuitive: it means that in order to uniquely recover the ground truth parameters, our training data must contain training examples that have more than $D_0$ observed features. This is different from some previous works (\cite{mohan2013graphical} for example), where complete case data must be available. 

Finally, given these new assumptions, it is easy to show that:
\begin{corollary}[Local identifiability]\label{cor:1} Assume that $p(X,Z|U) = p_{\epsilon}(X-f(Z))p_{T,\zeta}(Z|U)$ is the model parameterized according to Theorem \ref{th:1}. Assume that the assumptions in Theorem \ref{th:1} holds for $p(X|U)$. Additionally, assume that $f$ satisfies assumption \textbf{A5}. 

Then, consider the subset of variables, $X_o$. Then, $p(X_o|U)$ is $\sim_{A}$-identifiable on $(f_O,T,\zeta)$ for all $O$ that satisfies $|O|\geq D_0$, where $f_O$ is the entries from the output of $f$,  that corresponds to the index set $O$. 
\end{corollary}

\paragraph{Proof}: it is trivial to see that the assumptions 1, 3, and 4 in Theorem \ref{th:1} automatically holds regarding $p(X_o|U)$. $f_O$ is injective according to Assumption \textbf{A5}. Hence, $p(X_o|U)$ satisfies all the assumptions in Theorem \ref{th:1}, and $p(X_o|U)$ is $\sim_{A}$-identifiable on $(f_O,T,\zeta)$ for all $O$ that satisfies $|O|\geq D_0$. \qed

\paragraph{Remark} In practice,  \textbf{Assumption A5} is often satisfied. For example, consider the $f$ that is parameterized by the following MLP composite function:
\begin{equation}
    f(Z) = h(W\circ g(Z)) \label{eq:ivae_new_para}
\end{equation}
, where $g$ is a $D_0$ dimensional, injective multivariate function $g: \mathbb{R}^H \mapsto \mathbb{R}^{D_0}$, $h$ is some activation function $h:\mathbb{R} \mapsto \mathbb{R}$, and $W$ is a injective linear mapping $W: \mathbb{R}^{D_0} \mapsto \mathbb{R}^{D}$ represented by the matrix $W_{D_0 \times D}$, whose submatrices that consists of $|O| \geq D_0$ arbitrary selected columns are also injective. Note that this assumption for $W$ is not hard to fulfill: a randomly generated matrix (e.g., with element-wise i.i.d. Gaussian prior) satisfies this condition with probability 1.  To verify $f_O$ is injective for all $|O|\geq D_0$, notice that $f_O(Z) = h(W_O\circ g(Z))$, where $W_O$ is the output dimensions of $W$ that corresponds to the index set $O$. Since $W$ is injective and $|O|\geq D_0$, we have that $W_O$ is also injective, hence $f_O$ is also injective. 

\section{Consistency of estimation for GINA} \label{app:consistency} 

In \cite{khemakhem2020variational}, a result regarding the consistency of estimation for identifiable VAE. Similarly, it is trivial to show that similar result holds for GINA:

\begin{theorem}[Consistency of estimation]\label{th:consistency}

Assuming that

\begin{enumerate}
    \item $q_\phi(Z = z^k|X_o)$ is expressive enough to contain the true posterior $p_{\theta, \lambda}(Z|X_o)$, for all $X_o$, $\theta$ and $\lambda$. 
    \item The model in Section \ref{sec:practical_model} is correctly specified, and its parameters are estimated by maximizing  $\mathcal{L}_K(\theta,\lambda,\phi,X_o,R)$ w.r.t. $\theta$, $\lambda$, and $\phi$.
\end{enumerate}

Then, under perfect information (infinite samples from data), $\theta^*$ and $\lambda^*$ is recovered up to $\sim_A$. 
\end{theorem}

\paragraph{Proof} Since $q_\phi(Z = z^k|X_o)$ is expressive enough to contain the true posterior, $\mathcal{L}_K(\theta,\lambda,\phi,X_o,R)$ recovers the true likelihood function $\log p_{\theta,\lambda}(X_o,R)$ by simply maximizing $\phi$. Therefore, the problem of maximizing $\mathcal{L}_K(\theta,\lambda,\phi,X_o,R)$ is equivalent to maximum likelihood estimation problem. Therefore, since we assumed that the model is correctly specified, the consistency of MLE trivially implies the consistency of GINA model trained via maximizing $\mathcal{L}_K(\theta,\lambda,\phi,X_o,R)$.  \qed

\section{Active question selection}\label{app:eddi}
Suppose $X_o$ be the set of observed variables, that represents the correctness of student's response to questions that are presented to them. Then, in the problem of active question selection, we start with $O = \emptyset$, and we would like to decide which variable $X_i$ from $X_U$ to observe/query next, so that it will most likely provide the most valuable information for some target variable of interest, $X_\phi$;  Meanwhile, we should while make as few queries as possible. Once we have decided which $X_i$ to observed next, we will make query and add $i$ to $O$. This process is done by maximizing the information reward proposed by \cite{ma2018eddi}:
$$i^* = \argmax_{i \in U} R(i \mid X_O): = \mathbb{E}_{X_i \sim p(X_i|X_O)}
\mathbb{KL}\left[p(X_\phi | X_i,X_O) \,\|\, p(X_\phi | X_O)
\right].$$

In the Eedi dataset, as we do not have a specific target variable of interest, it is defined as $X_{\phi} = X_U$. In this case, $X_{\phi}$ could be ver high-dimensional, and direct estimation of $\mathbb{KL}\left[p(X_\phi | X_i,X_O) \,\|\, p(X_\phi | X_O)
\right].$ could be inefficient. In \cite{ma2018eddi}, a fast approximation has been proposed:
\begin{align*} 
R(i\mid X_o)  = & \mathbb{E} _{ X_i \sim p(X_i|X_o)} D_{KL}\left[ p(Z|X_i,X_o)||p( Z|X_o)\right] - \\ \nonumber
&\mathbb{E} _{X_{\phi},X_i \sim p(X_{\phi}, X_i|X_o)}  D_{KL}\left[ p(Z|X_{\phi},X_i,X_o)||p( Z|X_{\phi}, X_o) \right]. \\
\approx & \mathbb{E} _{ X_i \sim \hat{p}(X_i|X_o)} D_{KL}\left[ q(Z|X_i,X_o)||q( Z|X_o)\right] -\\ \nonumber
&\mathbb{E} _{X_{\phi},X_i \sim \hat{p}(X_{\phi}, X_i|X_o)}  D_{KL}\left[ q(Z|X_{\phi},X_i,X_o)||q( Z|X_{\phi}, X_o) \right].
\end{align*}
In this approximation, all calculation happens in the latent space of the model, hence we can make use of the learned inference net to efficeintly estimate $R(i\mid X_o)$.

\section{Additional results} \label{app:additional_results}

\subsection{Imputation results for synthetic datasets}
\begin{figure}[t]
    \centering
    \includegraphics[width=0.9\textwidth]{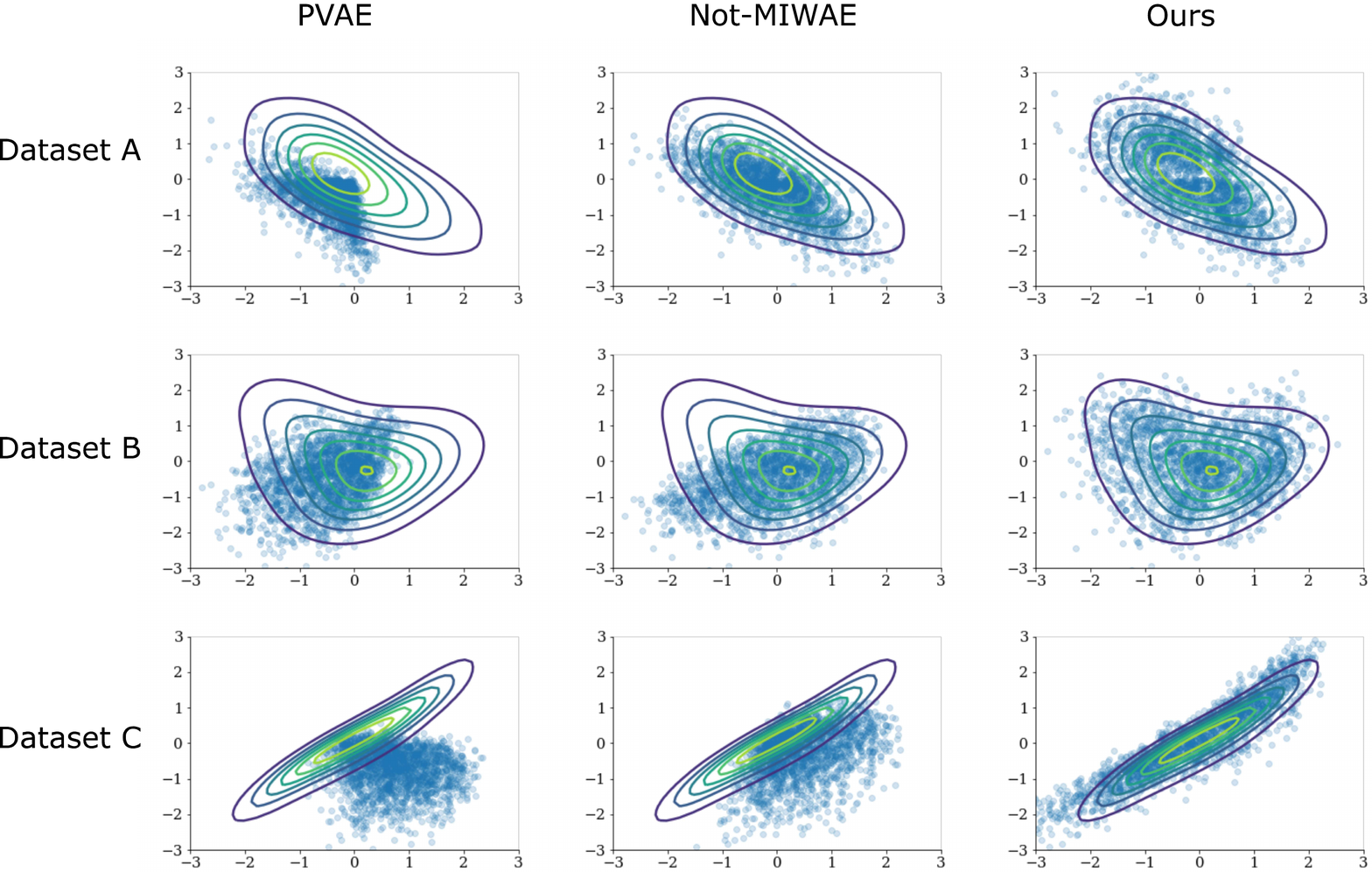}
    \caption{Visualization of imputed $X_2$ and $X_3$ from synthetic experiment. 
    \textbf{Row-wise (A-C)} plots for dataset A, B, and C, respectively; \textbf{Column-wise:}  PVAE imputed samples, Not-MIWAE imputed samples, and \name imputed samples, respectively. \textbf{Contour plot}: kernel density estimate of ground truth density of complete data;
    }
    \label{fig:synth_impute}
\end{figure}

In addition to the data generation samples visualized in Figure \ref{fig:synth}, we present the imputation results for synthetic datasets in Figure \ref{fig:synth_impute}. The procedure of generating the imputed samples are as follows. First, each model are trained on the randomly generated, partially observed synthetic dataset described in Section \ref{sec:exp_synth}. Once the models are trained, they are used to impute the missing data in the training set. For each training data, we draw exactly one sample from the (approximate) conditional distribution $p_
theta(X_u|X_o)$. Thus, we have ``complete'' version of the training set, one for each different model. Finally, we draw the scatter plot for each imputed training set, per dataset and per model. If the model is doing a good job recovering the ground truth distribution $p_\mathcal{D}(X)$ from training set, then its scatter plot should be close to the KDE estimate of the ground truth density of complete data. According to Figure \ref{fig:synth_impute}, the imputed distribution is similar to the generated distribution in Figure 
\ref{fig:synth}.

\end{document}